\documentclass[11pt]{article}

\usepackage[final]{acl}
\usepackage{xcolor}
\usepackage{times}
\usepackage{latexsym}
\usepackage{mdframed} 
\usepackage{tcolorbox} 
\definecolor{boxbg}{RGB}{245,245,245} 
\usepackage[T1]{fontenc}
\usepackage{tcolorbox}  
\usepackage{amsmath,amssymb}
\usepackage[utf8]{inputenc}

\usepackage{microtype}

\usepackage{inconsolata}

\usepackage{graphicx}

\usepackage{multirow}
\usepackage{longtable}
\usepackage{array}
\usepackage{geometry}
\usepackage{booktabs}
\usepackage{multirow}
\usepackage{array}
\usepackage{caption}
\usepackage[table]{xcolor} 
\usepackage{amsmath, graphicx, multirow, booktabs, float}
\usepackage{array}
\usepackage{makecell}
\usepackage{caption}
\usepackage{graphicx}
\usepackage{subcaption}

\newcolumntype{C}{>{\centering\arraybackslash}p{2.0cm}} 
\title{AutoSearch: Adaptive Search Depth for Efficient Agentic RAG via Reinforcement Learning}

\author{
Jingbo Sun\textsuperscript{1,2}\textsuperscript{*}\textsuperscript{\textdaggerdbl},
Wenyue Chong\textsuperscript{1,3}\textsuperscript{*}\textsuperscript{\textdaggerdbl},
Songjun Tu\textsuperscript{1,2}, 
Qichao Zhang\textsuperscript{1,2,3}\textsuperscript{\textdagger},
Yaocheng Zhang\textsuperscript{1,3}\textsuperscript{\textdaggerdbl}, \\
\textbf{
Jiajun Chai\textsuperscript{4}, 
Xiaohan Wang\textsuperscript{4}, 
Wei Lin\textsuperscript{4}, 
Guojun Yin\textsuperscript{4}, 
Dongbin Zhao\textsuperscript{1,2,3}} \\
\small
\textsuperscript{1}SKL-MAIS, Institute of Automation, Chinese Academy of Sciences\\
\small
\textsuperscript{2}School of Artificial Intelligence, University of Chinese Academy of Sciences \\
\small
\textsuperscript{3} School of Advanced Interdisciplinary Sciences, University of Chinese Academy of Sciences,
\textsuperscript{4}Meituan
}

\begin{document}
\maketitle

\let\thefootnote\relax\footnotetext{%
\hspace{-2.5em} 
* Equal contribution. † Corresponding authors. ‡ Intern work done at Meituan. This work is supported by the Strategic Priority Research Program of Chinese Academy of Sciences under Grant XDA0480302 and Meituan. Our code is available at \url{https://github.com/bofusun/AutoSearch}.
}

\begin{abstract}


Agentic retrieval-augmented generation (RAG) systems enable large language models (LLMs) to solve complex tasks through multi-step interaction with external retrieval tools.
However, such multi-step interaction often involves redundant search steps, incurring substantial computational cost and latency.
Prior work limits search depth (i.e., the number of search steps) to reduce cost, but this often leads to underexploration of complex questions.
To address this, we first investigate how search depth affects accuracy and find a minimal sufficient search depth that defines an accuracy-efficiency trade-off, jointly determined by question complexity and the agent’s capability.
Furthermore, we propose \textbf{AutoSearch}, a reinforcement learning (RL) framework that evaluates each search step via self-generated intermediate answers.
By a self-answering mechanism, AutoSearch identifies the minimal sufficient search depth and promotes efficient search by rewarding its attainment while penalizing over-searching.
In addition, reward mechanisms are introduced to stabilize search behavior and improve answer quality on complex questions.
Extensive experiments on multiple benchmarks show that AutoSearch achieves a superior accuracy-efficiency trade-off, alleviating over-searching while preserving search quality.

\end{abstract}

\section{Introduction}
Recently, LLMs~\cite{deepseekai2025deepseekr1incentivizingreasoningcapability, autothink, yue2026promoting, fu2025srft, yang2025qwen3, fu2025rlae} have demonstrated remarkable capabilities across complex and knowledge-intensive tasks.
Rather than relying solely on the internal knowledge of LLMs, RAG systems~\cite{jin2025search, song2025r1} leverage reinforcement learning to train LLMs as decision-making agents that autonomously invoke external retrieval tools, enabling effective handling of complex questions.
Agentic RAG systems typically operate in an autonomous loop of multi-turn reasoning and information retrieval, progressively acquiring evidence and culminating in a final response to complex questions.
From this perspective, prior agentic RAG research~\cite{gao2026unlocking} demonstrates that extended reasoning–retrieval trajectories are essential for solving complex questions.
However, increased search steps often incur over-searching, resulting in redundant retrievals and substantial computational cost and latency.

Recent studies~\cite{wu2025search, yang2026your} have explored RL signals to regulate the search behavior of agentic RAG systems and mitigate over-searching.
One line of research designs reward functions that explicitly penalize search depth or trajectory length~\cite{acting_less}.
Another approach~\cite{wu2025hiprag} penalizes redundant search steps based on internal model confidence or external large language model (LLM) judgments. While effective in reducing overall search frequency, such penalty-based methods may lead to insufficient search information for complex questions, potentially compromising final answer accuracy.
This motivates a key question:
\begin{tcolorbox}[
    colframe=black,    
    colback=boxbg,     
    boxrule=1.2pt,     
    arc=3pt,           
    boxsep=3pt,        
    left=5pt,
    right=5pt
]
Can agents adaptively adjust minimal search steps while preserving answer accuracy?
\end{tcolorbox}

\begin{figure}[t]
  \centering
  \includegraphics[width=\columnwidth]{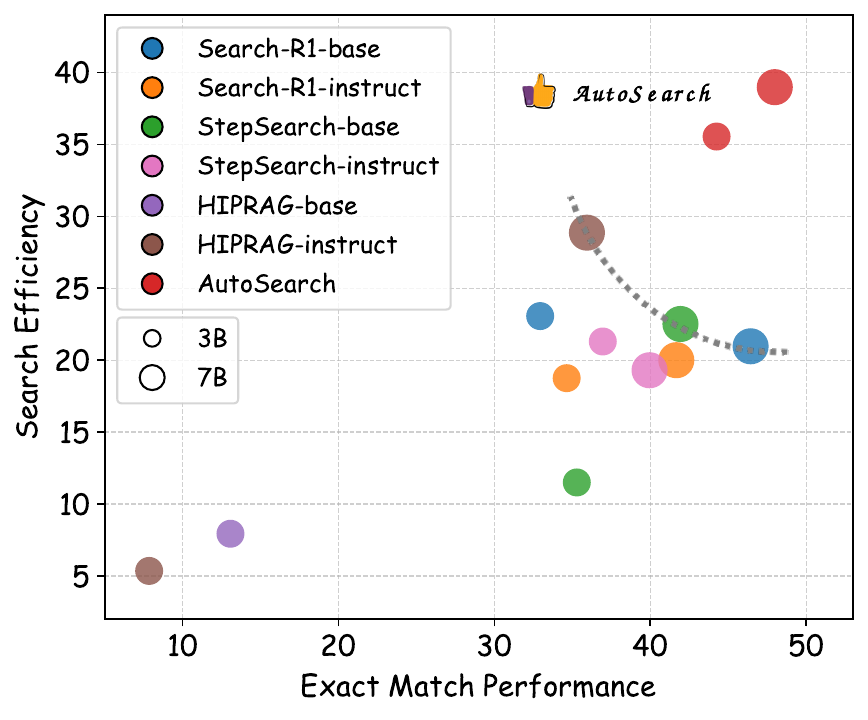} 
  \caption{Accuracy-efficiency Trade-off. AutoSearch attains leading EM with the highest search efficiency.}
  \label{fig:em_vs_se}
  \vspace{-.2cm}
\end{figure}

To address this question, we investigate how search depth affects answer accuracy and find a minimal sufficient search depth, jointly determined by question complexity and the agent's capability, that defines an accuracy-efficiency trade-off.  
Motivated by this, we propose \textbf{AutoSearch}, a self-decision-driven RL framework for agentic RAG systems.  
AutoSearch generates intermediate answers at each search step to identify the minimal sufficient search depth and encourages efficient search by rewarding attainment of this depth while penalizing over-searching.  
A base reward stabilizes the search process, while a search quality reward, derived from intermediate answers, further improves search quality on complex questions.  
With this self-adaptive framework, AutoSearch enables agents to autonomously adjust retrieval depth based on question complexity and their capability, achieving a principled accuracy-efficiency trade-off.
The main contributions are as follows:
\begin{itemize}
\item  We study how search depth affects answer accuracy and find that a minimal sufficient search depth, jointly determined by question complexity and the agent’s capability, establishes an accuracy-efficiency trade-off.
\vspace{-.3cm}
\item 
We propose a self-answer-driven RL framework that evaluates each search step by generating intermediate answers, training agents to adaptively adjust the minimal sufficient search depth while maintaining answer accuracy.
\vspace{-.3cm}
\item  Extensive experiments on single-hop and multi-hop QA tasks show that AutoSearch achieves a favorable trade-off between answer accuracy and search efficiency (Fig.~\ref{fig:em_vs_se}).
\end{itemize}

\vspace{-.1cm}
\section{Related Work}
\subsection{Agentic RAG System with RL}
Agentic RAG methods~\cite{gao2023retrieval, fan2024survey} have been proposed to enhance reasoning by integrating external knowledge retrieval.
Motivated by the effectiveness of RL~\cite{sun2026saliency, tu2025perception, sun2025unsupervised,lu2025equilibrium_neurips,liu2025videos}, recent studies~\cite{deepseekai2025deepseekr1incentivizingreasoningcapability,openai2024openaio1card} employ RL to train agents to strategically manage retrieval.
Search-R1~\cite{jin2025search} utilizes an RL framework with outcome-based rewards, enabling LLMs to interleave self-reasoning with real-time search.
R1-Searcher~\cite{song2025r1} proposes a two-stage RL method to improve the reasoning process for answer generation.
However, these RL-based approaches struggle with complex tasks due to sparse reward signals.
To mitigate this, recent works use process rewards to provide informative feedback in RL.
StepSearch~\cite{zheng2025stepsearch} introduces step-wise rewards that leverage information gain and penalize redundancy.
Atom-Searcher~\cite{deng2025atom} further decomposes reasoning into atomic thoughts and provides atomic-level process rewards.
Although prior approaches improve search performance, they overlook the issue of over-searching, leading to high computational cost.

\subsection{Efficient Agentic RAG}
To reduce redundant retrievals, recent studies design reward functions that penalize excessive retrieval. R1-Searcher++~\cite{song2025r1+} rewards the use of internal knowledge and caches retrieved content to prevent repetition, while another study~\cite{acting_less} proposes a trajectory-level reward that compares the current number of tool calls with an estimated optimal number to improve tool efficiency.
Other methods guide retrieval based on uncertainty or external judgment. Search Wisely~\cite{wu2025search} uses a confidence threshold to reward high-certainty search decisions. ReARTeR~\cite{sun2025rearter} employs a trustworthy process reward that combines a precise process reward model with an explanation-driven refinement model. HIPRAG~\cite{wu2025hiprag} uses an external LLM to supervise the necessity of each search step.
However, these approaches ~\cite{yue2026promoting, qian2025toolrl, shen2024smartcal} ignore how search depth should vary with problem complexity and model capability. In contrast, our method dynamically adjusts search depth accordingly, achieving an accuracy-efficiency trade-off.


\section{How Search Depth Affect Performance?}

\begin{figure*}[t]
    \centering
    \begin{minipage}[b]{0.47\textwidth}
        \centering
        \includegraphics[height=6.0cm, keepaspectratio]{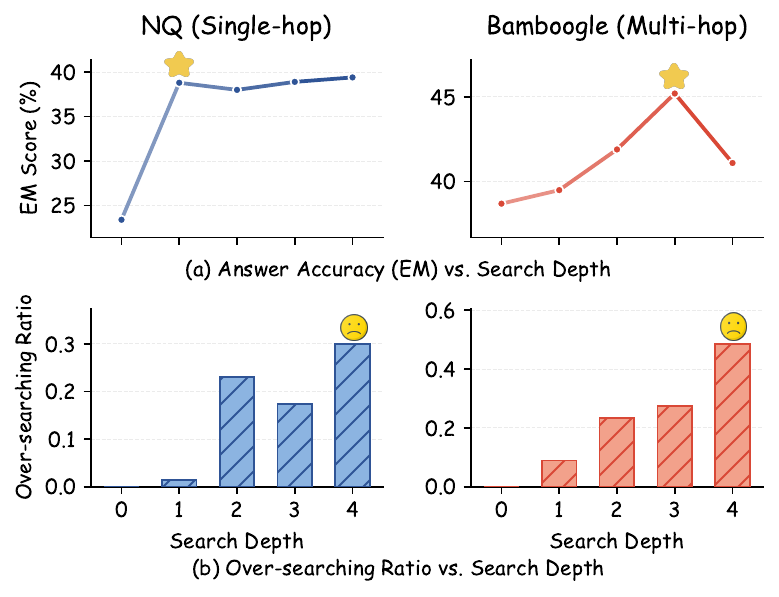}
    \end{minipage}%
    \hfill
    \begin{minipage}[b]{0.51\textwidth}
        \centering
        \includegraphics[height=6.0cm, width=1.005\linewidth, keepaspectratio]{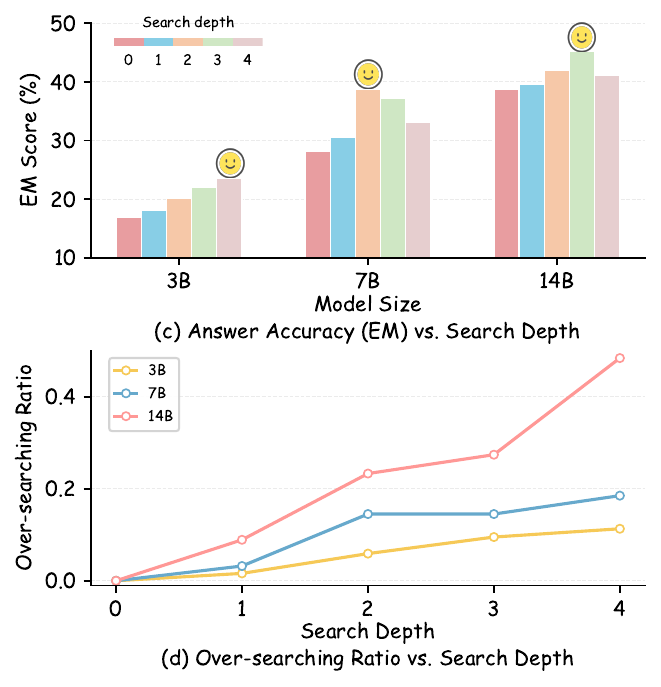}
    \end{minipage}
    \caption{
    Effect of search depth on accuracy and over-searching across dataset complexity and agent capability.}
    \label{fig:combined}
    \vspace{-.2cm}
\end{figure*}

\subsection{What is the Appropriate Search Depth?}
In agentic RAG systems, agents typically perform multi-step searches to acquire external knowledge. 
However, what is the appropriate search depth that balances accuracy and efficiency remains underexplored.
To investigate this, we analyze the impact of search depth on answer accuracy and over-searching behavior. 
Specifically, we prompt the Qwen2.5-14B-Instruct model to perform a fixed number of search steps for a given question. 
Experiments are conducted on the single-hop QA dataset Natural Questions (NQ) and the multi-hop QA dataset Bamboogle.
We measure answer accuracy using Exact Match (EM) and define an over-searching ratio as the proportion of samples where the agent could answer correctly before the final search step.
As shown in Fig.\ref{fig:combined}(a), NQ reaches near-optimal performance with a single search step, while additional steps provide little benefit. 
In contrast, Bamboogle requires up to three steps to peak, after which further searches reduce performance. 
Fig.~\ref{fig:combined}(b) further shows that exceeding the near-optimal depth leads to a notable increase in the over-searching ratio for both datasets, with the effect being more pronounced on the more complex Banboogle dataset.
These results indicate that \textbf{the appropriate search depth is task-dependent, with a minimal sufficient depth achieves near-optimal accuracy while mitigating over-searching.}

\subsection{What Affects Appropriate Search Depth?}

Different LLMs exhibit varying reasoning and retrieval capabilities, which in turn affect their answer accuracy in agentic RAG systems.
This raises a natural question: \emph{does the agent's capability also influence the appropriate search depth?}
To investigate this, we conduct controlled experiments on the Bamboogle dataset using Qwen-3B, Qwen-7B, and Qwen-14B models, each performing a range of search steps to answer the same set of questions.
As shown in Fig.~\ref{fig:combined}(c), the 3B model steadily improves with increasing search depth, achieving its peak performance at the final search step within the evaluated range.
In contrast, the 7B and 14B models reach their optimal performance at the second and third search steps, respectively, after which additional searches lead to performance degradation.
Fig.~\ref{fig:combined}(d) further shows that over-searching becomes increasingly pronounced as search depth grows, with stronger models exhibiting more pronounced degradation.
Taken together, these results from the previous and current subsections indicate that the minimal sufficient search depth, which balances accuracy and efficiency, is \textbf{jointly determined by question complexity and the agent’s capability}.

\section{Autosearch}

\begin{figure*}[t]
    \centering
        \centering
        \includegraphics[width=1.0\textwidth]
        {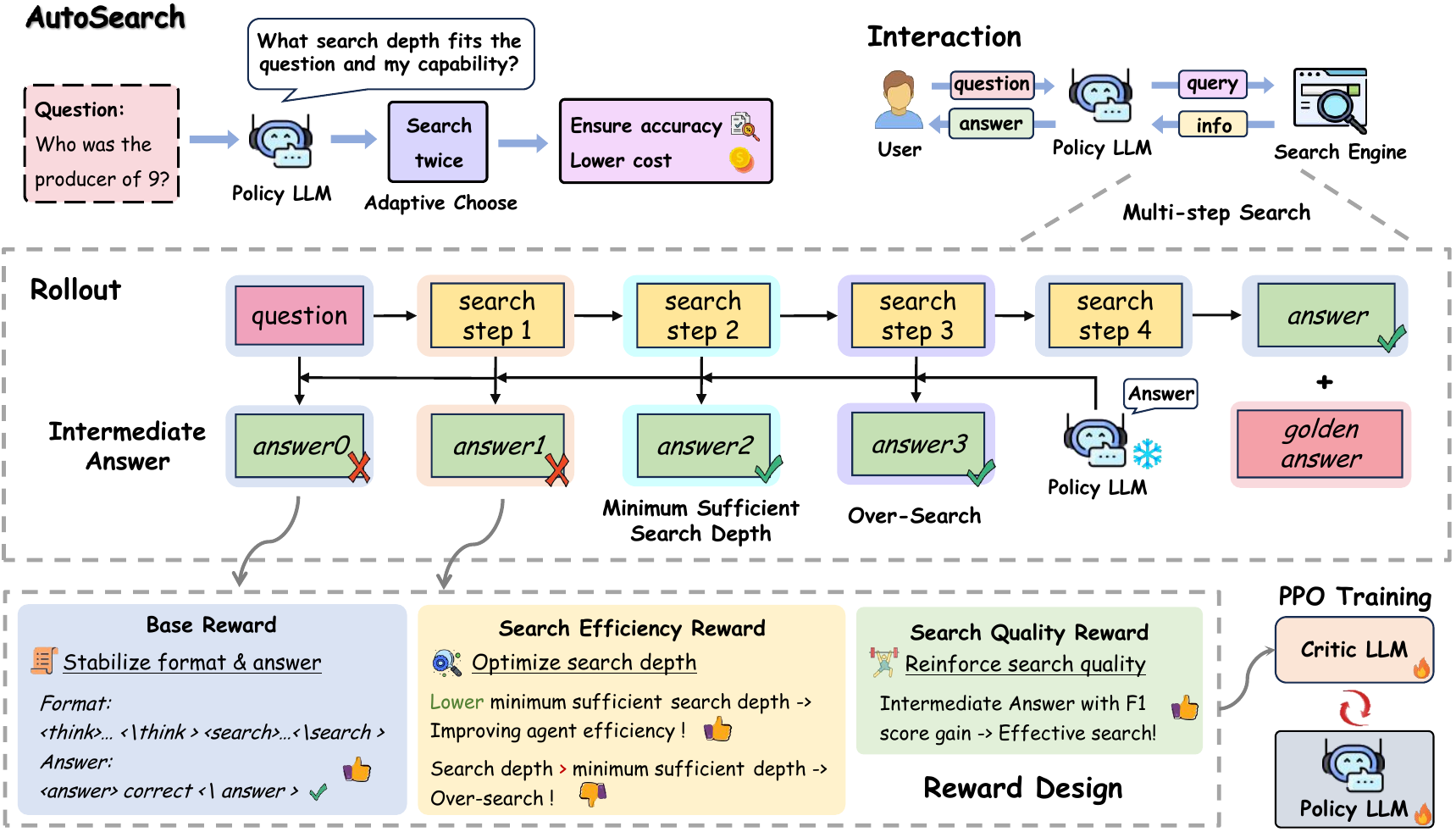}
    \caption{Overview of AutoSearch, which adaptively selects the search depth to trade off accuracy and efficiency.}
    \label{fig:overview}
    \vspace{-.2cm}
\end{figure*}

In this section, we present \textbf{AutoSearch}, a self-answer-driven RL framework for adaptive search in agentic RAG systems. 
Unlike prior approaches that separately optimize answer quality and search efficiency, AutoSearch enables the agent to jointly improve both with self-evaluative signals. 
Fig.\ref{fig:overview} provides an overview of AutoSearch, which incorporates three complementary reward signals:
(i) a \textbf{base reward} that regularizes search behavior,
(ii) a \textbf{search efficiency reward} that improves search efficiency while preserving answer quality, and
(iii) a \textbf{search quality reward} that enhances search quality.
The agent is trained with these rewards to balance search depth and search capability in an adaptive manner, accounting for both question complexity and its evolving answer capability.

\subsection{Base reward}
We introduce a \emph{base reward} to ensure stable multi-step retrieval and accurate answers, comprising a \emph{format reward} to enforce structured actions and an \emph{outcome reward} to encourage correct answers.
The format reward enforces a structured action format that separates reasoning, retrieval, and final answering. 
During retrieval steps, an action is considered invalid if it violates the prescribed format. 
Additionally, repeated retrieval of previously observed information is also treated as invalid, discouraging redundant search behavior.
At the final step, only the format of the produced answer is evaluated.
Rather than assigning a single reward to the entire trajectory, we compute the validity of each step independently to enable fine-grained credit assignment.
Formally, for the $i$-th question at step $t$, we define a binary indicator $\mathrm{valid}_{i,t} \in \{0,1\}$, indicating whether the action at step $t$ is well-formed and non-redundant.
The step-wise format reward is then defined as:
\begin{equation}
r_{i,t}^{\text{format}} =
\begin{cases}
0, & \mathrm{valid}_{i,t}=1,\ t < T, \\
-0.05, & \mathrm{valid}_{i,t}=0,\ t < T, \\
0.1, & \mathrm{valid}_{i,t}=1,\ t = T, \\
-0.5, & \mathrm{valid}_{i,t}=0,\ t = T,
\end{cases}
\end{equation}
where $T$ denotes the terminal step.
The outcome reward measures answer correctness using the Exact Match (EM) metric between the agent's predicted answer $a_{\text{pred}}$ and the ground-truth answer $a_{\text{gold}}$:
\begin{equation}
r^{\text{result}}_{i,t} =
\begin{cases}
\text{EM}(a_{i}^{\text{pred}}, a_{i}^{\text{gold}}), & t = T, \\
0, & t < T.
\end{cases}
\end{equation}
The overall base reward at step $t$ is computed as the sum of the format and outcome rewards:
\begin{equation}
r_{i,t}^{\text{base}} = r_{i,t}^{\text{format}} + r_{i,t}^{\text{result}}.
\end{equation}

\subsection{Search Efficiency Reward}
To improve search efficiency while maintaining answer quality, we propose a search efficiency reward that adaptively determines search depth based on the agent’s capability. 
To ensure answer quality, we generate intermediate answers at each retrieval step to identify the capability-aware optimal search depth, defined as the minimal number of retrieval steps required for the agent to answer correctly.  
Based on this depth, we design a reward function that encourages the agent to reduce its search depth while penalizing redundant retrievals, thereby improving search efficiency.


\paragraph{Intermediate Answer Generation.}
To estimate the optimal search depth for each trajectory based on current agent’s capability, the agent produces an intermediate answer at each retrieval step. 
For a multi-step search process on a given question $q_i$, at step $t$ the agent produces an intermediate answer based on all accumulated trajectory information:
\begin{equation}
x_{i,t} = \{ q_i, s_{i,0}, o_{i,0}, \ldots, s_{i,t}, o_{i,t} \},
\end{equation}
where $s_{i,t}$ and $o_{i,t}$ denote the search query and the retrieved observation at step $t$, respectively.
At each retrieval step $t$, the current policy $\pi_\theta$ is prompted using Prompt 1 to produce the intermediate answer:
\begin{equation}
a^{\text{inter}}_{i,t} = \pi_\theta(x_{i,t}),
\end{equation}
which serves as a self-evaluation signal guiding the assessment of both search depth and quality.

\begin{tcolorbox}[
  colframe=black,        
  colback=blue!30!gray!20,         
  coltitle=white,        
  colbacktitle=black, 
  fonttitle=\bfseries,   
  title={Prompt 1: Intermediate Answer Prompt},  
  boxrule=1pt,           
  left=10pt, right=10pt, top=5pt, bottom=5pt,  
  toptitle=3pt, bottomtitle=3pt,  
]

Answer the given question according to search trajectories, which consists of multiple reasoning, search calls, and retrieved information.
Important instructions: 

\textcolor{red!60}{(1) You must conduct reasoning inside <think> and </think> first.} 

\textcolor{red!60}{(2) After reasoning, output the final answer wrapped in <answer> and </answer>. }

For example: <think> Reasoning </think> <answer> Jaden Smith </answer>. 

Search trajectory: Question: \textcolor{red}{question}. Below are your previous reasoning, search calls, and retrieved information: \textcolor{blue}{trajectory}.

\end{tcolorbox}






\paragraph{Capability-aware Optimal Search Depth.}
Using the sequence of intermediate answers, we identify the capability-aware optimal search depth $t_c$.  
Specifically, $t_c$ is defined as the earliest step at which the intermediate answer $a^{\text{inter}}_{i,t}$ exactly matches the ground-truth answer $a_{\text{gold}}$:
\begin{equation}
t_c = \min \left\{ t \mid \text{EM}(a^{\text{inter}}_{i,t}, a_{\text{gold}}) = 1 \right\}.
\end{equation}
If the agent fails to answer correctly, we set $t_c=-1$.
This depth represents the minimal retrieval steps required for the agent to answer the question correctly, reflecting its capability for the given task.

\begin{table*}[t]
\centering
\caption{ \textbf{(Main Results)} EM, F1, Search Depth and Search Efficiency Comparison Across Methods.}
\label{tab:qa_main_metrics}
\small
\setlength{\tabcolsep}{4.0 pt}
\renewcommand{\arraystretch}{1.0}

\begin{tabular}{l *{14}{c}}
\toprule
\multirow{3}{*}{\textbf{Method}} &
\multicolumn{6}{c}{\textbf{General QA}} &
\multicolumn{8}{c}{\textbf{Multi-Hop QA}} \\
\cmidrule(lr){2-7} \cmidrule(lr){8-15}
&
\multicolumn{2}{c}{\textbf{NQ}} &
\multicolumn{2}{c}{\textbf{TriviaQA}} &
\multicolumn{2}{c}{\textbf{PopQA}} &
\multicolumn{2}{c}{\textbf{HotpotQA}} &
\multicolumn{2}{c}{\textbf{2Wiki}} &
\multicolumn{2}{c}{\textbf{Bamboogle}} &
\multicolumn{2}{c}{\textbf{Avg}} \\
\cmidrule(lr){2-3} \cmidrule(lr){4-5} \cmidrule(lr){6-7}
\cmidrule(lr){8-9} \cmidrule(lr){10-11} \cmidrule(lr){12-13} \cmidrule(lr){14-15}
& 
\scriptsize \textbf{EM $\uparrow$} & \scriptsize \textbf{F1 $\uparrow$} &
\scriptsize \textbf{EM $\uparrow$} & \scriptsize \textbf{F1 $\uparrow$} &
\scriptsize \textbf{EM $\uparrow$} & \scriptsize \textbf{F1 $\uparrow$} &
\scriptsize \textbf{EM $\uparrow$} & \scriptsize \textbf{F1 $\uparrow$} &
\scriptsize \textbf{EM $\uparrow$} & \scriptsize \textbf{F1 $\uparrow$} &
\scriptsize \textbf{EM $\uparrow$} & \scriptsize \textbf{F1 $\uparrow$} &
\scriptsize \textbf{EM $\uparrow$} & \scriptsize \textbf{F1 $\uparrow$} \\

\midrule


\rowcolor{green!10}
\multicolumn{15}{l}{\textit{\textbf{Base model: Qwen2.5-3b-Base/Instruct}}} \\

Search-R1-base           &\underline{36.0}& \underline{44.9}& 51.3& 59.0 &\underline{37.0} &\underline{41.3} & 27.9& 36.7& 29.3& 34.6 & 16.1& 25.9 &32.9 &40.4\\
Search-R1-instruct           &32.2 &42.0 &\underline{52.4} &\underline{60.3} &35.7 &41.2 &30.3 &39.6 &32.2 &38.0 &25.0 &36.9 &34.6 &43.0 \\
StepSearch-base         &30.0 &39.9 &47.3 &55.0 &32.2 &37.1 &32.1 &42.6 &\underline{37.1} &\underline{42.8} &33.0 &\underline{41.9} &35.3 &43.2\\
StepSearch-instruct         &33.9 &43.5 &51.1 &59.3 &34.9 &40.2 &\underline{33.2} &\underline{44.2} &34.7 &41.5 &\textbf{33.9} &\textbf{44.7} &\underline{37.0} &\underline{45.6} \\
HIPRAG-base              &11.0 &20.2 &23.4 &34.1 &14.9 &23.0 &10.3 &19.1 &8.30 &16.2 &10.5 &16.8 &13.1 &21.6\\
HIPRAG-instruct              &3.20 &9.50 &14.5 &24.2 &6.60 &13.7 &10.1 &18.4 &8.70 &18.6 &4.00 &11.9 &7.85 &16.1\\
\rowcolor{yellow!15}
AutoSearch     &\textbf{43.7} &\textbf{52.2} &\textbf{61.7} &\textbf{69.0} &\textbf{46.0} &\textbf{50.1} &\textbf{39.4} &\textbf{50.5} &\textbf{41.6} &\textbf{47.7} &\underline{33.1} &\textbf{44.7} &\textbf{44.3} &\textbf{52.4}\\

\midrule
\multicolumn{1}{c}{} &
\makecell{\scriptsize \textbf{SD $\downarrow$}} &
\makecell{\scriptsize \textbf{SE $\uparrow$}} &
\makecell{\scriptsize \textbf{SD $\downarrow$}} &
\makecell{\scriptsize \textbf{SE $\uparrow$}} &
\makecell{\scriptsize \textbf{SD $\downarrow$}} &
\makecell{\scriptsize \textbf{SE $\uparrow$}} &
\makecell{\scriptsize \textbf{SD $\downarrow$}} &
\makecell{\scriptsize \textbf{SE $\uparrow$}} &
\makecell{\scriptsize \textbf{SD $\downarrow$}} &
\makecell{\scriptsize \textbf{SE $\uparrow$}} &
\makecell{\scriptsize \textbf{SD $\downarrow$}} &
\makecell{\scriptsize \textbf{SE $\uparrow$}} &
\makecell{\scriptsize \textbf{SD $\downarrow$}} &
\makecell{\scriptsize \textbf{SE $\uparrow$}} \\

Search-R1-base           &\underline{1.23} &\underline{29.4} &\underline{1.29} &\underline{39.9} &\underline{1.35} &\underline{27.4} &1.63 &\underline{17.2} &1.95 &15.0 &\underline{1.69} &9.50 &1.52 &\underline{23.1} \\
Search-R1-instruct           &1.76 &18.3 &1.62 &32.3 &1.72 &20.7 &2.09 &14.5 &2.29 &14.1 &1.98 &12.6 &1.91 &18.8 \\
StepSearch-base         &2.95 &10.2 &2.90 &16.3 &2.94 &10.9 &3.19 &10.1 &3.51 &10.6 &3.03 &10.9 &3.09 &11.5\\
StepSearch-instruct         &1.46 &23.3 &1.57 &32.7 &1.45 &24.1 &2.05 &16.2 &2.29 &\underline{15.1} &2.07 &\underline{16.4} &1.82 &21.3\\
HIPRAG-base              &1.59 &6.92 &1.60 &14.6 &1.68 &8.88 &1.66 &6.22 &\underline{1.70} &4.87 &1.72 &6.11 &1.66 &7.93\\
HIPRAG-instruct              &1.28 &2.50 &1.32 &11.0 &\underline{1.35} &4.88 &\underline{1.58} &6.40 &1.84 &4.74 &\textbf{1.56} &2.57 &\underline{1.49} &5.35\\
\rowcolor{yellow!15}
AutoSearch     &\textbf{1.03} &\textbf{42.4} &\textbf{1.05} &\textbf{58.7} &\textbf{1.07} &\textbf{42.9} &\textbf{1.49} &\textbf{26.5} &\textbf{1.69} &\textbf{24.7} &1.83 &\textbf{18.1} &\textbf{1.36} &\textbf{35.6}\\

\midrule

\rowcolor{green!10}
\multicolumn{15}{l}{\textit{\textbf{Base model: Qwen2.5-7b-Base/Instruct}}} \\

\multicolumn{1}{c}{} &
\makecell{\scriptsize \textbf{EM $\uparrow$}} &
\makecell{\scriptsize \textbf{F1 $\uparrow$}} &
\makecell{\scriptsize \textbf{EM $\uparrow$}} &
\makecell{\scriptsize \textbf{F1 $\uparrow$}} &
\makecell{\scriptsize \textbf{EM $\uparrow$}} &
\makecell{\scriptsize \textbf{F1 $\uparrow$}} &
\makecell{\scriptsize \textbf{EM $\uparrow$}} &
\makecell{\scriptsize \textbf{F1 $\uparrow$}} &
\makecell{\scriptsize \textbf{EM $\uparrow$}} &
\makecell{\scriptsize \textbf{F1 $\uparrow$}} &
\makecell{\scriptsize \textbf{EM $\uparrow$}} &
\makecell{\scriptsize \textbf{F1 $\uparrow$}} &
\makecell{\scriptsize \textbf{EM $\uparrow$}} &
\makecell{\scriptsize \textbf{F1 $\uparrow$}} \\

Search-R1-base          &\textbf{47.6} &\textbf{56.3} &\underline{63.7} &\underline{71.4} &\underline{45.7} &\underline{49.8} &\textbf{43.7} &\textbf{55.4} &38.7 &44.8 &41.1 &\underline{53.4} &\underline{46.8} &\underline{55.2}\\
Search-R1-instruct           &\underline{40.2} &48.5 &61.0 &68.4 &40.5 &45.4 &38.7 &49.0 &32.5 &38.1 &37.1 &49.6 &41.7 &49.8\\
StepSearch-base         &37.7 &47.9 &54.7 &63.3 &38.6 &44.0 &37.0 &48.6 &\underline{40.9} &\underline{47.3} &\textbf{42.7} &51.9 &41.9 &50.5\\
StepSearch-instruct         &34.2 &44.5 &51.6 &59.9 &34.9 &40.1 &38.9 &50.4 &39.8 &46.3 &40.3 &52.4 &40.0 &48.9\\
HIPRAG-instruct              &30.6 &42.0 &55.8 &65.2 &34.2 &42.4 &30.5 &42.7 &29.8 &38.3 &34.7 &47.8 &35.9 &46.4\\
\rowcolor{yellow!15}
AutoSearch    &\textbf{47.6} &\underline{56.0} &\textbf{65.8} &\textbf{73.1} &\textbf{46.3} &\textbf{50.3} &\underline{42.7} &\underline{54.2} &\textbf{43.6} &\textbf{50.3} &\underline{41.9} &\textbf{54.9} &\textbf{48.0} &\textbf{56.5}\\

\midrule
\multicolumn{1}{c}{} &
\makecell{\scriptsize \textbf{SD $\downarrow$}} &
\makecell{\scriptsize \textbf{SE $\uparrow$}} &
\makecell{\scriptsize \textbf{SD $\downarrow$}} &
\makecell{\scriptsize \textbf{SE $\uparrow$}} &
\makecell{\scriptsize \textbf{SD $\downarrow$}} &
\makecell{\scriptsize \textbf{SE $\uparrow$}} &
\makecell{\scriptsize \textbf{SD $\downarrow$}} &
\makecell{\scriptsize \textbf{SE $\uparrow$}} &
\makecell{\scriptsize \textbf{SD $\downarrow$}} &
\makecell{\scriptsize \textbf{SE $\uparrow$}} &
\makecell{\scriptsize \textbf{SD $\downarrow$}} &
\makecell{\scriptsize \textbf{SE $\uparrow$}} &
\makecell{\scriptsize \textbf{SD $\downarrow$}} &
\makecell{\scriptsize \textbf{SE $\uparrow$}} \\

Search-R1-base           &1.98 &24.0 &1.86 &34.2 &1.95 &23.4 &2.65 &16.5 &3.15 &12.3 &2.56 &16.1 &2.36 &21.1\\
Search-R1-instruct           &1.95 &20.7 &1.74 &35.0 &1.79 &22.6 &2.43 &15.9 &3.08 &10.5 &2.42 &15.3 &2.24 &20.0\\
StepSearch-base         &1.68 &22.4 &1.73 &31.6 &1.57 &24.7 &2.07 &17.8 &2.29 &\underline{17.9} &2.07 &20.7 &1.90 &22.5\\
StepSearch-instruct         &1.94 &17.6 &1.99 &25.9 &1.98 &17.6 &2.10 &18.6 &2.46 &16.2 &2.02 &19.9 &2.08 &19.3\\
HIPRAG-instruct              &\underline{1.07} &\underline{28.5} &\textbf{0.97} &\underline{57.8} &\underline{1.16} &\underline{29.5} &\underline{1.55} &\underline{19.7} &\underline{2.15} &13.8 &\textbf{1.45} &\underline{23.9} &\underline{1.39} &\underline{28.9}\\
\rowcolor{yellow!15}
AutoSearch    &\textbf{1.05} &\textbf{45.3} &\underline{1.04} &\textbf{63.3} &\textbf{1.09} &\textbf{40.8} &\textbf{1.34} &\textbf{31.2} &\textbf{1.58} &\textbf{25.9} &\underline{1.53} &\textbf{27.4} &\textbf{1.27} &\textbf{39.0} \\

\bottomrule
\end{tabular}
\end{table*}

\begin{figure}[ht]
    \centering
    \includegraphics[width=0.48\textwidth]{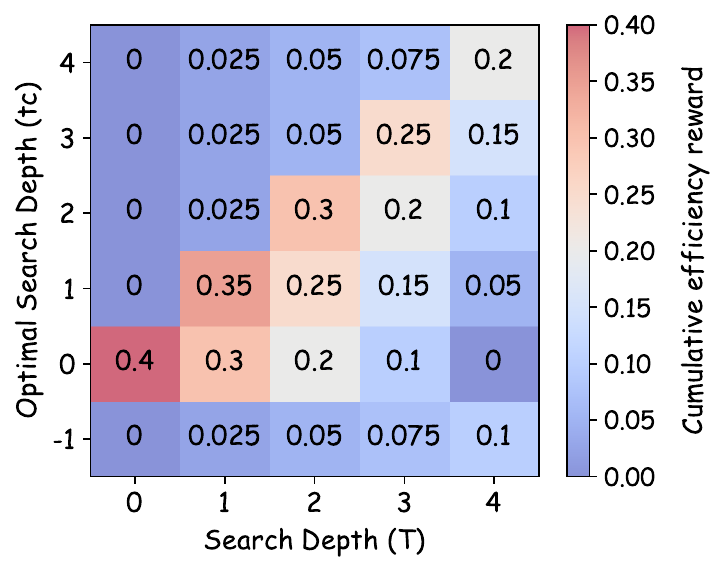} 
    \caption{Cumulative reward versus search depth. The cumulative reward peaks at optimal search depth and decreases for insufficient searching and over-searching.}
    \label{fig:figure3}
    \vspace{-.4cm}
\end{figure}

\paragraph{Search Efficiency Reward.}  
We categorize each retrieval step into three types:
\textbf{(i) Under-search:} If the agent fails to answer the question correctly ($t_c > T$), all steps in the trajectory are considered under-search. Each step is assigned a small positive reward to encourage further information acquisition.  
\textbf{(ii) Effective search:} If the agent answers correctly at step $t_c$, the steps preceding $t_c$ are considered effective search. Each of these steps receives a positive reward, with higher rewards for shorter $t_c$ to encourage more efficient search.  
\textbf{(iii) Over-searching:} Steps after $t_c$ are considered over-searching and are penalized to discourage redundant retrieval.  
Formally, the search efficiency reward at step $t$ is defined as
\begin{equation}
r^{\text{efficient}}_{i,t} =
\begin{cases}
0.025, & t_{c}<0, \\
\frac{0.4}{t_{c} + \epsilon} - 0.05, & 0<t \le t_c< T-1, \\
-0.1, &  0<t_c < t < T-1,
\end{cases}
\end{equation}
where $\epsilon$ is a small constant for numerical stability.  
Fig.~\ref{fig:figure3} shows that, under a fixed agent capability, cumulative Efficiency reward is maximized when search terminates at the capability-aware optimal depth.
Both insufficient searching and over-searching lead to lower rewards, encouraging the agent to learn an optimal search depth rather than simply minimizing retrieval steps.

\subsection{Search Quality Reward}
To improve search quality in complex tasks, we introduce a search quality reward based on intermediate answers.
We quantify search quality by computing the F1 score between the intermediate answer and the ground-truth answer.
Let $PN$, $RN$, and $IN$ denote the number of tokens in the intermediate answer, the ground-truth answer, and their overlap, respectively.
The F1 score is defined as:
\begin{equation}
\mathrm{F1}(a^{\text{inter}}_{i,t}, a^{\text{gold}}_{i}) = \frac{2 \cdot IN}{PN + RN}.
\end{equation}
We define the search quality reward as the gain over the best F1 score achieved in previous steps:
\begin{equation}
\normalsize 
r^{\text{quality}}_{i,t} =
\mathrm{F1}(a^{\text{inter}}_{i,t}, a^{\text{gold}}_{i})
- \max_{t' < t} \mathrm{F1}(a^{\text{inter}}_{i,t'}, a^{\text{gold}}_{i}).
\end{equation}
This reward explicitly measures the marginal contribution of each retrieval step to answer quality, encouraging the agent to perform searches that provide meaningful and non-redundant information.

\subsection{RL Training}
We train the agent using a composite reward that jointly encourages answer correctness and search efficiency.
Specifically, the overall reward at retrieval step $t$ is defined as:
\begin{equation}
r^{\text{overall}}_{i,t}
=
r^{\text{base}}_{i,t}
+
r^{\text{efficient}}_{i,t}
+
r^{\text{quality}}_{i,t},
\end{equation}
where the base reward ensures answer correctness, the efficiency reward encourages effective search trajectories, and the quality reward promotes informative retrieval steps. 
We adopt Proximal Policy Optimization (PPO) as our RL backbone, while masking external observations. We apply a token-level mask $I(o_t)$ to exclude retrieved tokens from the training loss. The training objective is: 
\begin{equation}
\small
\mathcal{J}_{\text{PPO}}(\theta)
=
\mathbb{E}_{\substack{(q,a)\sim\mathcal{D} \\ o \sim \pi_{\text{old}}(\cdot|q)}}
\left[
\frac{1}{\sum_{t=1}^{|o|} I(o_t)}
\sum_{\substack{t=1 \\ I(o_t)=1}}^{|o_i|}
L_{t}
\right],
\end{equation}
where $o$ denotes the search trajectory for question $q$, $|o|$ denotes the trajectory length, and loss $L_{t}$ is:
\begin{equation}
\small
L_{t}
=
\min\!\left(
r_{t}\hat{A}_{t},
\,
\text{clip}(r_{t},\,1-\varepsilon,\,1+\varepsilon)\hat{A}_{t}
\right),
\end{equation}
where $r_t$ is importance sampling ratio and $\hat{A}_t$ denotes the advantage estimate computed using GAE.

\section{Experiment}

\subsection{Setup}

\paragraph{Datasets} We evaluate AutoSearch on six widely used QA datasets, divided into two categories. 
(1) General QA: NQ~\cite{kwiatkowski2019natural}, TriviaQA~\cite{joshi2017triviaqa}, and PopQA~\cite{mallen2022not}; 
(2) Multi-hop QA: HotpotQA~\cite{yang2018hotpotqa}, 2WikiMultiHopQA~\cite{ho2020constructing}, and Bamboogle~\cite{press2023measuring}. 
These datasets cover a broad spectrum of scenarios, allowing a comprehensive evaluation of AutoSearch.

\paragraph{Baselines} We compare AutoSearch against several RL-based search baselines, including Search-R1~\cite{jin2025search}, StepSearch~\cite{zheng2025stepsearch}, and HiPRAG~\cite{wu2025hiprag}.




\paragraph{Evaluation Metrics}
We assess answer quality using the canonical \textbf{Exact Match (EM)} and word-level \textbf{F1} scores. 
To evaluate search cost and efficiency, we report \textbf{Search Depth (SD)}, which records the number of search steps, and \textbf{Search Efficiency (SE)} ($\text{SE} = \text{EM} / \text{SD}$), quantifying the accuracy gained per search step. 
We also introduce the \textbf{Over-Searching Ratio (OSR)}, the proportion of samples correctly answered before the final search step, to quantify over-searching.

\begin{table}[t]
\centering
\caption{Over-Searching Ratio (OSR\% $\downarrow$) Comparison.}
\label{tab:qa_main_metrics_ec_only}

\footnotesize
\setlength{\tabcolsep}{0.5pt}
\newcolumntype{C}{>{\centering\arraybackslash}p{0.11\linewidth}}

\begin{tabular}{l *{6}{C}}
\toprule
\multirow{2}{*}{\textbf{Method}} &
\multicolumn{3}{c}{\textbf{General QA}} &
\multicolumn{3}{c}{\textbf{Multi-Hop QA}} \\
\cmidrule(lr){2-4} \cmidrule(lr){5-7}
&
\textbf{NQ} &
\textbf{Tri.} &
\textbf{Pop.} &
\textbf{Hot.} &
\textbf{2Wi.} &
\textbf{Bam.} \\
\midrule
\rowcolor{purple!10}
\multicolumn{7}{l}{\textit{\textbf{Base model: Qwen2.5-3b-Base/Instruct}}} \\

Search-R1-base      &3.50 &2.10 &5.60 &9.20 &11.1 &6.50 \\
Search-R1-instruct  &8.70 &3.90 &8.90 &13.9 &11.3 &16.9 \\
StepSearch-base     &25.2 &12.2 &27.7 &33.9 &32.4 &37.1 \\
StepSearch-instruct &5.70 &4.30 &4.90 &18.1 &12.7 &15.3 \\
HIPRAG-base         &2.70 &1.80 &5.50 &2.60 &\textbf{0.80} &3.20 \\
HIPRAG-instruct     &0.70 &0.80 &1.80 &1.70 &1.40 &3.20 \\
\rowcolor{orange!10}
AutoSearch          &\textbf{0.00} &\textbf{0.10} &\textbf{0.00} &\textbf{0.90} &2.60 &\textbf{2.40} \\

\midrule
\rowcolor{purple!10}
\multicolumn{7}{l}{\textit{\textbf{Base model: Qwen2.5-7b-Base/Instruct}}} \\

Search-R1-base      &13.0 &6.50 &11.4 &32.4 &32.3 &30.6 \\
Search-R1-instruct  &9.30 &3.90 &6.80 &23.4 &27.2 &29.8 \\
StepSearch-base     &12.2 &7.10 &12.1 &23.8 &24.5 &29.0 \\
StepSearch-instruct &20.8 &10.9 &26.6 &27.3 &26.4 &32.3 \\
HIPRAG-instruct     &4.50 &2.20 &7.90 &12.5 &18.9 &11.3 \\
\rowcolor{orange!10}
AutoSearch          &\textbf{1.83} &\textbf{0.51} &\textbf{0.50} &\textbf{5.78} &\textbf{8.52} &\textbf{10.1} \\

\bottomrule
\end{tabular}
\end{table}

\paragraph{Training and Evaluation}
We conduct experiments on two models from the Qwen series: Qwen-2.5-3B-Base and Qwen-2.5-7B-Base. 
Following Search-R1, we merge the training sets of NQ and HotpotQA dataset to form training corpus. 
During training, the retrieval module used the 2018 Wikipedia snapshot~\cite{karpukhin2020dense} with the E5 retriever~\cite{wang2022text}, uniformly sampling $k=3$ documents.
To ensure fair and reproducible comparisons, we independently reproduce the results of baseline methods.



\subsection{Main Results}

\noindent\textbf{AutoSearch Achieves Superior Performance.} 
Table~\ref{tab:qa_main_metrics} reports the performance of AutoSearch and baseline methods across six QA datasets. 
Across both in-domain and out-of-domain settings, as well as single-hop and multi-hop QA tasks, AutoSearch consistently outperforms the baselines, achieving higher EM and F1 scores and demonstrating robust accuracy across diverse scenarios.

\noindent\textbf{AutoSearch Balances Accuracy and Efficiency.}
The search depth results show that Search-R1 and StepSearch achieve strong answer accuracy but rely on deeper searches, resulting in higher computational cost.
In contrast, HiPRAG adopts fewer search depth on most tasks, which reduces search cost but leads to noticeably lower accuracy.
Compared to baselines, AutoSearch consistently achieves higher accuracy with fewer search steps across most tasks, demonstrating balanced trade-off between answer accuracy and search efficiency.

\noindent\textbf{AutoSearch Adaptively Adjusts Search Depth.}  
AutoSearch does not simply minimize search depth. On single-hop datasets, it maintains relatively shallow searches, while on multi-hop datasets, it performs noticeably more search steps.  
The search efficiency metric shows that, compared to baseline methods, AutoSearch achieves higher efficiency across most datasets, indicating its ability to adaptively adjust search depth based on question complexity and the agent’s capability.

\noindent\textbf{AutoSearch Effectively Suppresses Over-Searching.}  
Table~\ref{tab:qa_main_metrics_ec_only} shows that Search-R1 and StepSearch exhibit high over-searching ratios, particularly on multi-hop tasks, indicating substantial redundant searches in prior methods.  
While HiPRAG reduces over-searching by leveraging external models, this comes at the cost of lower answer accuracy.  
In contrast, AutoSearch effectively suppresses over-searching while maintaining strong performance, leveraging the agent's own capability to adaptively control search depth.

\begin{table*}[t]
\centering
\caption{Ablation study on search reward components in AutoSearch using Qwen2.5-3B base model.}
\label{tab:ablation_search_rewards}
\small
\setlength{\tabcolsep}{2.5pt}
\renewcommand{\arraystretch}{0.9}


\begin{tabular}{l *{14}{c}}
\toprule
\multirow{3}{*}{\textbf{Method}} &
\multicolumn{6}{c}{\textbf{General QA}} &
\multicolumn{8}{c}{\textbf{Multi-Hop QA}} \\
\cmidrule(lr){2-7} \cmidrule(lr){8-15}
&
\multicolumn{2}{c}{\textbf{NQ}} &
\multicolumn{2}{c}{\textbf{TriviaQA}} &
\multicolumn{2}{c}{\textbf{PopQA}} &
\multicolumn{2}{c}{\textbf{HotpotQA}} &
\multicolumn{2}{c}{\textbf{2Wiki}} &
\multicolumn{2}{c}{\textbf{Bamboogle}} &
\multicolumn{2}{c}{\textbf{Avg}} \\

\cmidrule(lr){2-3} \cmidrule(lr){4-5} \cmidrule(lr){6-7}
\cmidrule(lr){8-9} \cmidrule(lr){10-11} \cmidrule(lr){12-13}
\cmidrule(lr){14-15}
& 
\scriptsize \textbf{EM $\uparrow$} & \scriptsize \textbf{SD $\downarrow$} &
\scriptsize \textbf{EM $\uparrow$} & \scriptsize \textbf{SD $\downarrow$} &
\scriptsize \textbf{EM $\uparrow$} & \scriptsize \textbf{SD $\downarrow$} &
\scriptsize \textbf{EM $\uparrow$} & \scriptsize \textbf{SD $\downarrow$} &
\scriptsize \textbf{EM $\uparrow$} & \scriptsize \textbf{SD $\downarrow$} &
\scriptsize \textbf{EM $\uparrow$} & \scriptsize \textbf{SD $\downarrow$} &
\scriptsize \textbf{EM $\uparrow$} & \scriptsize \textbf{SD $\downarrow$} \\

\midrule
\rowcolor{blue!10}
\multicolumn{15}{l}{\textit{\textbf{Base model: Qwen2.5-3b-Base}}} \\

AutoSearch 
& 43.7 & 1.03 
& \textbf{61.7} & 1.05 
& \textbf{46.0} & 1.07 
& \textbf{39.4} & 1.49 
& \textbf{41.6} & 1.69 
& 33.1 & 1.83
& \textbf{44.3} & 1.36 \\

w/o Base Reward  
& 40.4 & 1.36 
& 57.7 & 1.37 
& 44.2 & 1.38 
& 34.2 & 1.79 
& 37.2 & 2.02 
& 31.2 & 2.02
& 40.8 & 1.66 \\

w/o Search Efficiency Reward 
& \textbf{44.2} & 3.24 
& 61.5 & 3.22 
& 43.3 & 3.34 
& 36.6 & 3.21 
& 40.4 & 3.38 
& \textbf{33.8} & 3.30
& 43.3 & 3.28 \\

w/o Search Quality Reward            
& 40.2 & \textbf{1.01} 
& 58.6 & \textbf{1.03} 
& 42.9 & \textbf{1.01} 
& 33.2 & \textbf{1.09} 
& 36.1 & \textbf{1.22} 
& 30.8 & \textbf{1.25}
& 40.3 & \textbf{1.10} \\

\bottomrule
\end{tabular}
\vspace{-.3cm}
\end{table*}

\subsection{Ablation Study}  
We evaluate the contribution of the three reward mechanisms via ablation experiments on six QA datasets.  
Table~\ref{tab:ablation_search_rewards} shows that removing the base reward reduces accuracy and slightly increases search depth, highlighting its role in stabilizing the search process. Removing the search efficiency reward decreases accuracy and significantly increases search depth, indicating its importance for the accuracy-efficiency trade-off. Removing the search quality reward lowers accuracy and noticeably reduces search depth, showing that it encourages more informative search steps.  
Overall, these results confirm that all three reward components are crucial for balancing answer accuracy and search efficiency.

\subsection{Further Analysis}
\begin{figure}[t]
  \centering
  \includegraphics[width=0.9\columnwidth]{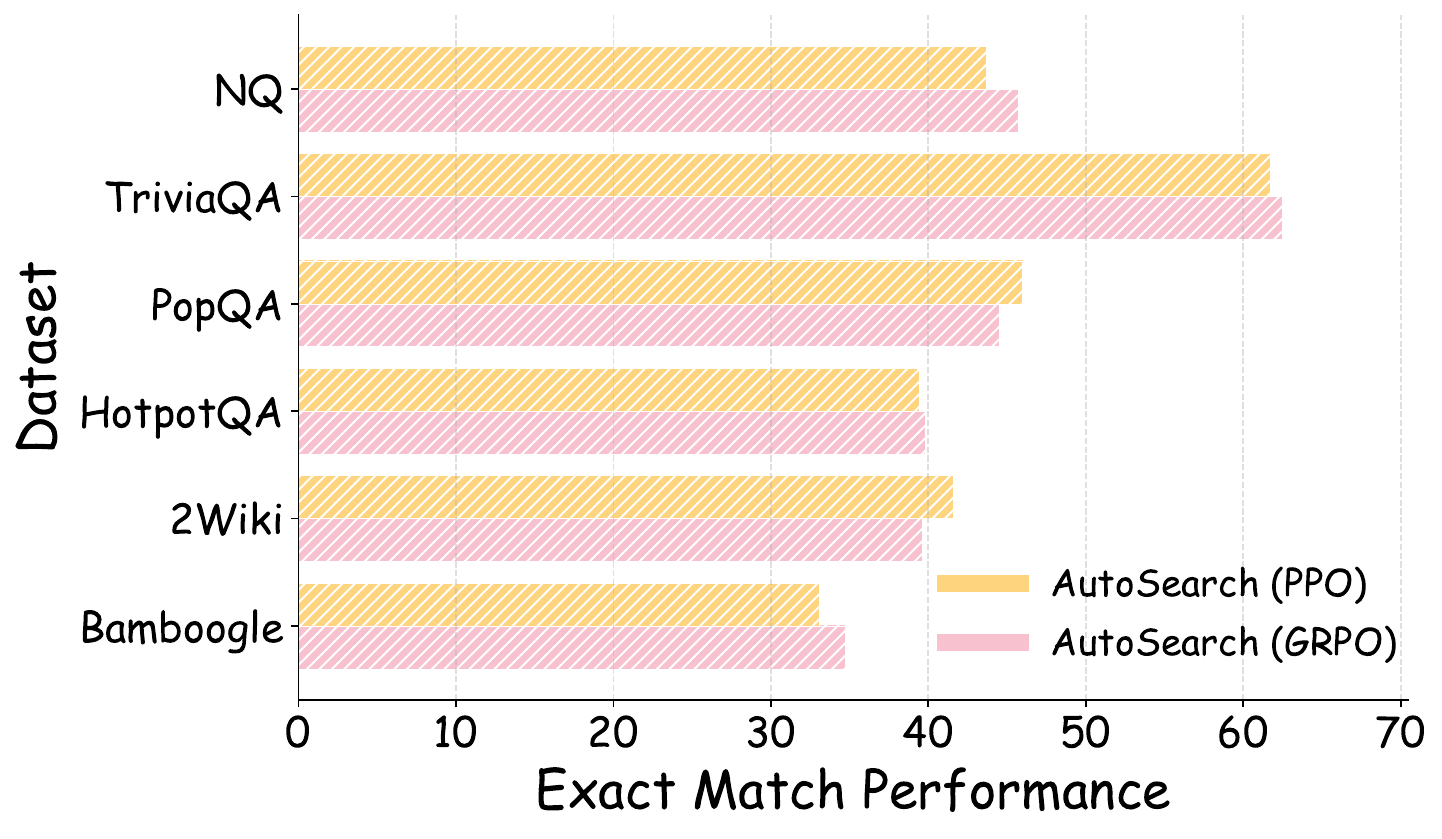} 
  \caption{The performance results(EM) of AutoSearch with PPO and GRPO on six datasets.}
  \label{fig:ppo_vs_grpo}
  \vspace{-.4cm}
\end{figure}

\noindent\textbf{Outcome-Based RL for AutoSearch.}
AutoSearch is primarily trained using step-wise rewards, which evaluate each search step.  
A natural question is whether outcome-based RL methods, such as GRPO, which optimize for final answer quality, can also be effective for AutoSearch.
To investigate this, we treat AutoSearch's cumulative reward as the outcome reward and apply GRPO for optimization.  
Figure~\ref{fig:ppo_vs_grpo} compares the performance of AutoSearch trained with PPO versus GRPO.  
Surprisingly, GRPO achieves performance comparable to PPO, indicating the generality of AutoSearch.




\paragraph{Training Dynamic of AutoSearch.}
As shown in Figure~6, we analyze the training dynamics of AutoSearch compared to Search-R1 and StepSearch.

\begin{figure}[t]
    \centering
    \includegraphics[width=0.48\textwidth]{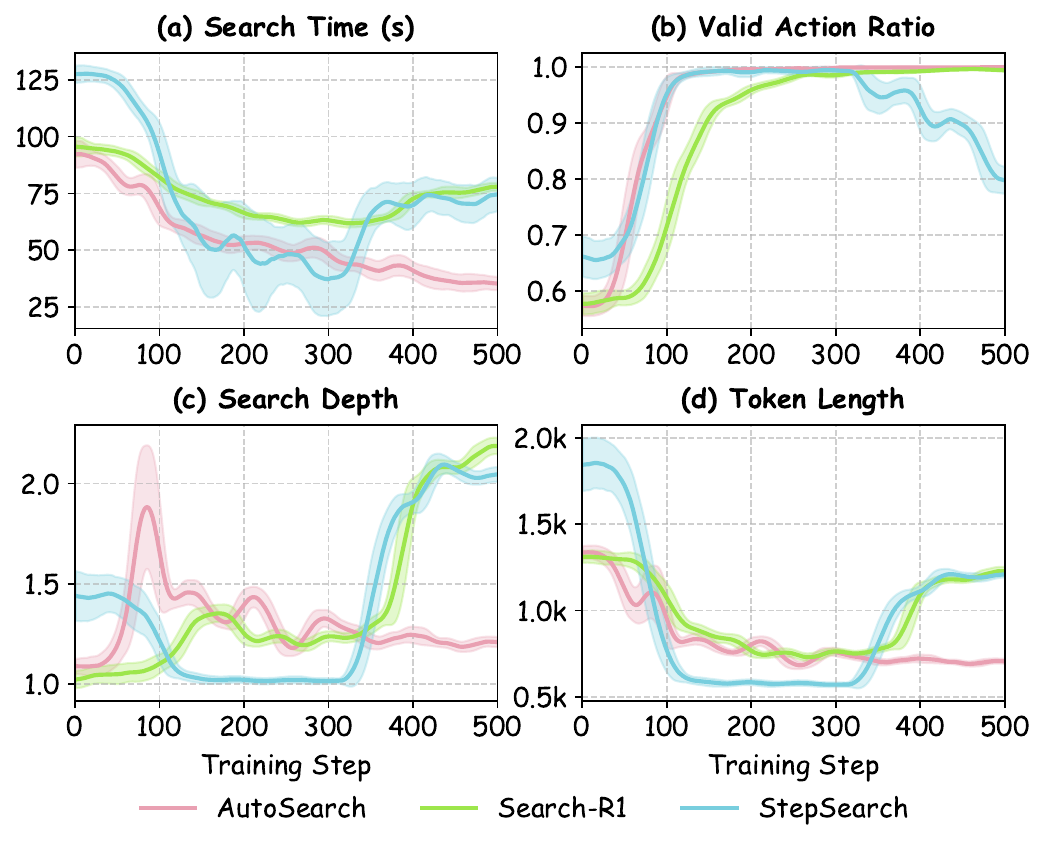} 
    \caption{Training dynamics of three models.}
    \label{fig:four-figures}
    \vspace{-.4cm}
\end{figure}

\noindent\textbf{Search Time.} Fig.\ref{fig:four-figures}(a) shows that while both Search-R1 and StepSearch initially reduce search time during early training, they later stabilize at relatively high values. In contrast, AutoSearch gradually decreases search time throughout training, achieving lower inference latency.  

\noindent\textbf{Valid Actions Ratio.} As illustrated in Fig.\ref{fig:four-figures}(b), all methods converge to valid actions. However, Search-R1 converges slowly, and StepSearch exhibits instability in later training. AutoSearch achieves the fastest convergence and maintains a high valid actions ratio, reflecting stable search.  

\noindent\textbf{Search Depth.} Fig.\ref{fig:four-figures}(c) shows that Search-R1 and StepSearch keep shallow search early but substantially increase search depth later. In contrast, AutoSearch briefly explores deeper search and then converges to a lower, stable depth, demonstrating higher efficiency.

\noindent\textbf{Token Length.} As shown in Figure~\ref{fig:four-figures}(d), Search-R1 and StepSearch initially generate short token length, but increases with deeper search, indicating higher resource consumption. In contrast, AutoSearch gradually reduces token length while maintaining accuracy, achieving a better balance between efficiency and answer quality.

\section{Conclusion}
In this work, we propose \textbf{AutoSearch}, a RL framework that adaptively selects the search depth based on question complexity and the agent’s capability. 
By analyzing how search depth affects answer performance, we identify a minimal sufficient depth that balances accuracy and efficiency. 
AutoSearch leverages intermediate answers at each step to adaptively reach this depth, ensuring efficient search without compromising accuracy. 
Further reward mechanisms are employed to stabilize the search process and improve answer quality on complex questions. 
Empirical results on multiple benchmarks demonstrate that AutoSearch achieves a favorable accuracy-efficiency trade-off.

\section*{Limitations}
Although AutoSearch achieves an accuracy-efficiency trade-off by adaptively selecting search depth, this study focuses on relatively low maximum search steps. Investigating the balance between search depth and accuracy under a broader range of search depths remains an interesting direction for future work.




\bibliography{custom}

\appendix

\clearpage

\section{Expanded Related Work}
\begin{table*}[htbp]
\centering
\caption{Ablation Study of Reward Mechanisms on Model Performance with F1 and OSR (\%) Metrics}
\label{tab:ablation_search_rewards1}
\small
\setlength{\tabcolsep}{2.5pt}
\renewcommand{\arraystretch}{1.1}

\begin{tabular}{l *{12}{c}}
\toprule
\multirow{3}{*}{\textbf{Method}} &
\multicolumn{6}{c}{\textbf{General QA}} &
\multicolumn{6}{c}{\textbf{Multi-Hop QA}} \\
\cmidrule(lr){2-7} \cmidrule(lr){8-13}
&
\multicolumn{2}{c}{\textbf{NQ}} &
\multicolumn{2}{c}{\textbf{TriviaQA}} &
\multicolumn{2}{c}{\textbf{PopQA}} &
\multicolumn{2}{c}{\textbf{HotpotQA}} &
\multicolumn{2}{c}{\textbf{2Wiki}} &
\multicolumn{2}{c}{\textbf{Bamboogle}} \\
\cmidrule(lr){2-3} \cmidrule(lr){4-5} \cmidrule(lr){6-7}
\cmidrule(lr){8-9} \cmidrule(lr){10-11} \cmidrule(lr){12-13}
& 
\scriptsize \textbf{F1 $\uparrow$} & \scriptsize \textbf{OSR $\downarrow$} &
\scriptsize \textbf{F1 $\uparrow$} & \scriptsize \textbf{OSR $\downarrow$} &
\scriptsize \textbf{F1 $\uparrow$} & \scriptsize \textbf{OSR $\downarrow$} &
\scriptsize \textbf{F1 $\uparrow$} & \scriptsize \textbf{OSR $\downarrow$} &
\scriptsize \textbf{F1 $\uparrow$} & \scriptsize \textbf{OSR $\downarrow$} &
\scriptsize \textbf{F1 $\uparrow$} & \scriptsize \textbf{OSR $\downarrow$} \\

\midrule
\rowcolor{blue!10}
\multicolumn{13}{l}{\textit{\textbf{Base model: Qwen2.5-3b-Base}}} \\

AutoSearch  &52.2 &1.03 &\textbf{69.0} &1.05 &\textbf{50.1} &1.07 &\textbf{50.5} &1.49 &\textbf{47.7} &1.69 &44.7 &1.83 \\
w/o Base Reward   &49.3 &1.16 &65.2 &0.99 &48.5 &2.05 &44.5 &4.54 &43.0 &5.82 &43.0 &1.61 \\
w/o Search Efficiency Reward  &\textbf{53.7} &2.09 &68.2 &11.4 &47.4 &26.5 &46.8 &23.1 &46.1 &15.8 &\textbf{45.3} &19.4 \\
w/o Search Quality Reward             &50.4 &\textbf{0.00} &65.8 &\textbf{0.02} &47.1 &\textbf{0.07} &43.2 &\textbf{0.43} &42.8 &\textbf{0.63} &42.7 &\textbf{0.42} \\

\bottomrule
\end{tabular}
\end{table*}

\begin{table*}[htbp]
\centering
\caption{Performance of AutoSearch with PPO and GRPO on Different Datasets}
\label{tab:qa_main_metrics_grpo}
\small
\setlength{\tabcolsep}{4.0pt}
\renewcommand{\arraystretch}{1.2}

\begin{tabular}{l *{12}{c}}
\toprule
\multirow{3}{*}{\textbf{Method}} &
\multicolumn{6}{c}{\textbf{General QA}} &
\multicolumn{6}{c}{\textbf{Multi-Hop QA}} \\
\cmidrule(lr){2-7} \cmidrule(lr){8-13}
&
\multicolumn{2}{c}{\textbf{NQ}} &
\multicolumn{2}{c}{\textbf{TriviaQA}} &
\multicolumn{2}{c}{\textbf{PopQA}} &
\multicolumn{2}{c}{\textbf{HotpotQA}} &
\multicolumn{2}{c}{\textbf{2Wiki}} &
\multicolumn{2}{c}{\textbf{Bamboogle}} \\
\cmidrule(lr){2-3} \cmidrule(lr){4-5} \cmidrule(lr){6-7}
\cmidrule(lr){8-9} \cmidrule(lr){10-11} \cmidrule(lr){12-13}
& 
\scriptsize \textbf{EM $\uparrow$} & \scriptsize \textbf{F1 $\uparrow$} &
\scriptsize \textbf{EM $\uparrow$} & \scriptsize \textbf{F1 $\uparrow$} &
\scriptsize \textbf{EM $\uparrow$} & \scriptsize \textbf{F1 $\uparrow$} &
\scriptsize \textbf{EM $\uparrow$} & \scriptsize \textbf{F1 $\uparrow$} &
\scriptsize \textbf{EM $\uparrow$} & \scriptsize \textbf{F1 $\uparrow$} &
\scriptsize \textbf{EM $\uparrow$} & \scriptsize \textbf{F1 $\uparrow$} \\

\midrule


\rowcolor{blue!10}
\multicolumn{13}{l}{\textit{\textbf{Base model: Qwen2.5-3b-Base/Instruct}}} \\

AutoSearch(PPO)          &43.7 &52.2 &61.7 &69.0 &\textbf{46.0} &\textbf{50.1} &39.4 &50.5 &\textbf{41.6} &\textbf{47.7} &33.1 &44.7 \\
AutoSearch(GRPO)         &\textbf{45.7} &\textbf{54.3} &\textbf{62.5} &\textbf{69.6} &44.5 &48.4 &\textbf{39.8} &\textbf{50.6} &39.6 &46.1 &\textbf{34.7} &\textbf{46.1} \\

\midrule
\multicolumn{1}{c}{} &
\makecell{\scriptsize \textbf{SD $\downarrow$}} &
\makecell{\scriptsize \textbf{SE $\uparrow$}} &
\makecell{\scriptsize \textbf{SD $\downarrow$}} &
\makecell{\scriptsize \textbf{SE $\uparrow$}} &
\makecell{\scriptsize \textbf{SD $\downarrow$}} &
\makecell{\scriptsize \textbf{SE $\uparrow$}} &
\makecell{\scriptsize \textbf{SD $\downarrow$}} &
\makecell{\scriptsize \textbf{SE $\uparrow$}} &
\makecell{\scriptsize \textbf{SD $\downarrow$}} &
\makecell{\scriptsize \textbf{SE $\uparrow$}} &
\makecell{\scriptsize \textbf{SD $\downarrow$}} &
\makecell{\scriptsize \textbf{SE $\uparrow$}} \\

AutoSearch(PPO)          &\textbf{1.03} &42.4 &1.05 &58.7 &\textbf{1.07} &\textbf{42.9} &1.49 &26.5 &1.69 &\textbf{24.7} &1.83 &18.1\\
AutoSearch(GRPO)         &\textbf{1.03} &\textbf{44.4} &\textbf{1.04} &\textbf{60.1} &1.08 &41.2 &\textbf{1.41} &\textbf{28.2} &\textbf{1.62} &24.4 &\textbf{1.61} &\textbf{21.6} \\

\midrule
\end{tabular}
\end{table*}
\subsection{Agentic RAG System with RL}
Large language models (LLMs) face challenges such as outdated knowledge and factual inaccuracies, making it difficult to rely on them for up-to-date or contextually correct responses. To mitigate these limitations, Agentic Retrieval-Augmented Generation (Agentic RAG) systems have emerged as an effective solution. These methods enhance the reasoning capabilities of LLMs by integrating external knowledge retrieval directly into their reasoning process~\cite{ gao2023retrieval, fan2024survey}. While early works in this area~\cite{trivedi2023interleaving, yu2022generate, shi2024replug}primarily relied on prompt engineering to enable agents to invoke search engines, they often struggled with ensuring that these agents could effectively use and interact with external tools.

In contrast, recent studies have begun utilizing reinforcement learning (RL) to train agents to strategically manage the retrieval process. For instance, Search-R1~\cite{jin2025search}introduces a reinforcement learning framework that enables LLMs to dynamically interleave self-reasoning with search, improving both the accuracy and efficiency of responses. R1-Searcher~\cite{song2025r1}proposes a two-stage RL method, where the first stage trains the agent to use external retrieval effectively, and the second stage integrates retrieval seamlessly into the reasoning process for more accurate answers. While these methods have demonstrated success in simpler tasks, they face limitations when applied to complex, multi-hop question answering (QA) tasks, due to sparse and delayed reward signals. This often leads to inefficiencies in utilizing the retrieved knowledge and may negatively impact performance on more challenging queries.

\subsection{Advancements in RL-Based Agentic RAG Systems}
Building upon the initial successes of Agentic RAG systems, recent studies have introduced novel methods to improve the performance of RL-based agents. To address the issue of sparse rewards, which can hinder the agent's learning process, several approaches have been proposed. For example, StepSearch~\cite{zheng2025stepsearch}introduces step-wise rewards that aim to provide more informative feedback by considering information gain and penalizing redundant retrieval steps. Atom-Searcher~\cite{deng2025atom}takes this further by decomposing reasoning tasks into smaller, atomic thoughts, enabling more detailed process rewards at the atomic level. This approach enhances the agent's ability to reason more efficiently. Despite these improvements, RL-based approaches often struggle with balancing the trade-off between search cost and reasoning quality, particularly when dealing with complex, multi-hop tasks where the retrieval process may need to be more dynamic and adaptive.

\subsection{Efficient Agentic RAG Systems}

To address the problem of inefficient search behavior, several studies have focused on designing reward functions that can reduce unnecessary retrievals. R1-Searcher++~\cite{song2025r1+}offers a solution by rewarding the effective use of internal knowledge and memorizing previously retrieved content to avoid repetition. This helps reduce the need for repeated searches, improving efficiency. Similarly, studies such as~\cite{acting_less}propose trajectory-level rewards that compare the number of retrieval steps taken by the agent to an estimated optimal number, further improving search efficiency. Other methods, such as Search Wisely~\cite{wu2025search}, guide retrieval decisions based on the agent's confidence, rewarding high-certainty search actions. While these techniques reduce redundant retrieval, they often overlook the dynamic nature of retrieval depth, which should vary depending on task complexity and the capabilities of the model. In contrast, our method dynamically adjusts the retrieval depth, which allows it to better manage trade-offs between performance and efficiency.

\section{Detailed Illustration of Our Method}
\begin{figure*}[htbp]
    \centering
        \centering
        \includegraphics[width=1.0\textwidth]
        {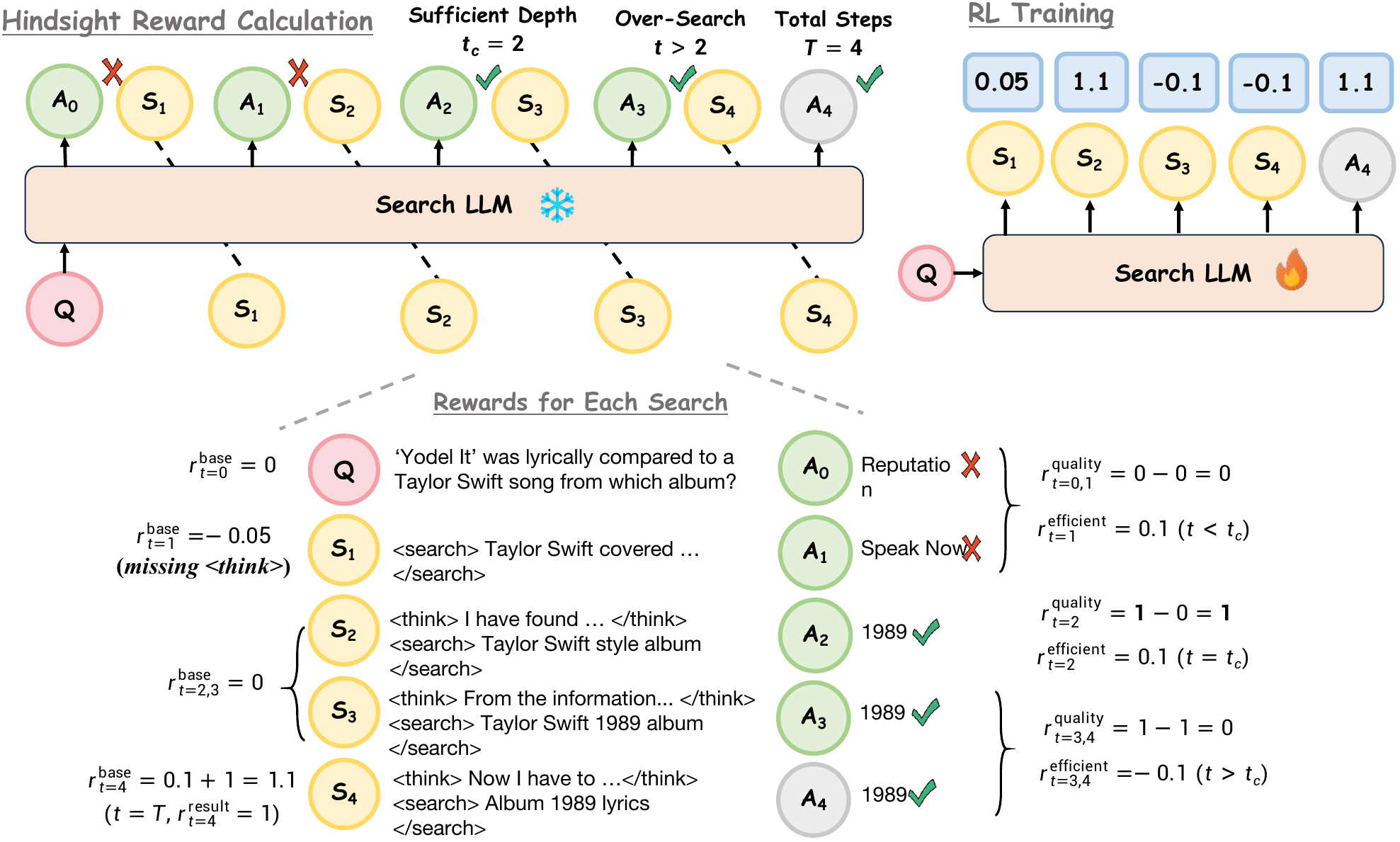}
    \caption{Illustrative Example of the Proposed Method}
    \label{fig:main2}
    \vspace{-.2cm}
\end{figure*}

Figure~\ref{fig:main2} presents a concrete example illustrating how our method assigns differentiated reward signals to insufficient and excessive search behaviors during training.  Through this carefully designed reward mechanism, the model learns to adaptively balance search adequacy and search efficiency while maintaining answer correctness.

\section{Experiment Setups}
\paragraph{Training Details.}
The training process consists of a total of 1,005 optimization steps. The policy network (actor) is optimized with a learning rate of 1$\times$10$^{-6}$,
while the value network (critic) uses a higher learning rate of 1$\times$10$^{-5}$. Warm-up scheduling is applied to both networks to stabilize early-stage training, with warm-up ratios set to 0.285 and 0.015 of the total training steps for the
actor and critic, respectively.

All experiments are conducted on a single compute node equipped with eight
NVIDIA H20 GPUs.
A total batch size of 512 is used and divided into PPO mini-batches of size 256.
To balance memory efficiency and computational throughput, the micro-batch sizes
are set to 64 for the actor and 8 for the critic.

\paragraph{Data and Context Length.}
Training data is sampled from the mixed nq and hotpotqa train dataset. The total token length aggregated across all turns is capped at 4,096 tokens
to ensure compatibility with the model's maximum context window.

\section{Ablation Study on Search Rewards}
We conducted an ablation study in Table\ref{tab:ablation_search_rewards1} to analyze the impact of different reward mechanisms on model performance. In addition to the EM and SD metrics discussed in the main text, we also further analyzed F1 and OSR. The results show that the base reward helps regularize the search behavior, preventing redundant steps; the search efficiency reward significantly improves the model's search efficiency, especially in complex multi-hop reasoning tasks; and the search quality reward effectively enhances the quality of the answers. By combining these three reward mechanisms, the model achieves a better balance between search depth and quality, thus improving both search efficiency and accuracy.

\section{Comparison of PPO and GRPO in AutoSearch}

In this table\ref{tab:qa_main_metrics_grpo}, we present the results of AutoSearch under PPO and GRPO training, evaluated using EM, F1, SD, and SE. PPO and GRPO exhibit task-dependent differences in AutoSearch, primarily in how they balance answer quality (EM/F1), search depth (SD), and search efficiency (SE). While both methods achieve comparable performance across datasets, GRPO tends to yield slightly higher EM/F1 on NQ and Bamboogle but lower scores on PopQA and 2Wiki. In terms of search behavior, GRPO generally uses similar or slightly fewer search steps than PPO—most notably on Bamboogle—and achieves higher SE on TriviaQA and Bamboogle, though slightly lower SE on PopQA and 2Wiki. These differences suggest that the two algorithms employ distinct policy optimization strategies, leading to varying trade-offs between exploration efficiency and answer accuracy depending on the dataset characteristics.

\section{Prompts}

As shown in Figure~\ref{fig:prompt-0}-~\ref{fig:prompt-4}, we provide a series of standardized prompt templates that enforce a fixed number of retrieval steps, ranging from 0 to 4. For the k-step retrieval setting (k = 0, 1, 2, 3, 4), each prompt explicitly instructs the model to perform exactly k search operations and to output the final answer concisely within <answer> and </answer>. This structured prompting strategy ensures strict experimental control, enabling a systematic evaluation of how varying retrieval depths affect performance on both single-hopand multi-hop question benchmarks.

\section{Case Study}

To demonstrate the effectiveness and efficiency of AutoSearch, we present qualitative analyses across five representative cases, comparing its reasoning and retrieval behavior against three strong baselines: Search-R1, StepSearch, and HiPRAG.

In Case 1 (Figure~\ref{fig:Search-R1 vs AutoSearch}), Search-R1 fails to retrieve relevant evidence in its initial query about the genre of Suits, triggering a cascade of errors that misclassifies the show as a “TV series and sitcom.” In contrast, AutoSearch immediately formulates a precise search intent and correctly identifies Suits as a “legal drama.”

In Case 2 (Figure~\ref{fig:StepSearch vs AutoSearch}), when tasked with classifying Katie Sagona’s acting category, StepSearch starts with an overly broad search and must sift through irrelevant results—such as other individuals named Sagona. Although it eventually concludes she is an “actress,” it misses the crucial nuance that she is specifically a “child actor.” AutoSearch, by comparison, rapidly zeroes in on the key concept—“child actor”—and delivers the exact ground-truth answer with fewer steps and higher precision.

In Case 3 (Figure~\ref{fig:StepSearch vs AutoSearch1}),StepSearch adopts a strictly decomposed, step-by-step strategy, separately querying the countries of Umm Al-Tuyour, Hama Governorate, and Sharqliyya. However, it exhibits clear deficiencies during the information integration stage. On the one hand, it misinterprets retrieved evidence (e.g., incorrectly identifying Umm Al-Tuyour as being located in Saudi Arabia); on the other hand, it fails to align its final response with the required yes/no question format, instead directly outputting the country name “Syria.” As a result, both the answer form and semantic correctness are compromised. In contrast, AutoSearch performs a holistic retrieval centered on the original question, using a single, more targeted query to simultaneously gather key information about all relevant locations. Through joint reasoning over the retrieved evidence, it correctly determines that all three locations are situated in Syria and ultimately provides the correct response in the expected format by answering “Yes.”

In Case 4 (Figure~\ref{fig:HIPRAG vs AutoSearch}), HiPRAG exhibits a tendency toward over-elaboration and procedural redundancy. Faced with the question regarding the public status of Ural Federal University and California State Polytechnic University, Pomona, HiPRAG correctly identifies the entities but unnecessarily fragments the process into two distinct steps. In the first step, it retrieves information and reaches a valid conclusion, yet it immediately initiates a second search and reasoning step to "verify" what it has already found. This leads to a repetitive response that restates the same information without adding value. AutoSearch, in contrast, demonstrates greater efficiency by directly interpreting the retrieved evidence—that both institutions are public universities—and outputs the correct answer in a single, concise reasoning step, avoiding the circular validation loop seen in HiPRAG.

In Case 5 (Figure~\ref{fig:HIPRAG vs AutoSearch1}), HiPRAG exhibits a failure in reasoning chain integration and evidence utilization. Faced with the question regarding the region of the city of Gualeguaychu, HiPRAG successfully executes a multi-step retrieval process—first identifying that the city is in the province of Entre Ríos, and then retrieving documents stating that Entre Ríos is in the Mesopotamia region. However, despite explicitly finding the correct region in the second step, HiPRAG fails to incorporate this evidence into its final conclusion. Instead, it hallucinates an incorrect answer ("central region"), likely due to a misinterpretation of the province's location or a failure to synthesize the findings. In contrast, Our Method demonstrates robust logical coherence; it follows the same necessary path of reasoning from city to province to region, but crucially, it correctly identifies and outputs "Mesopotamia" as the final answer, effectively utilizing the retrieved evidence.

These case studies underscore AutoSearch’s ability to adapt its search strategy dynamically based on query semantics and retrieved evidence that avoiding both under-reasoning and over-reasoning, thereby achieving superior performance in complex open-domain QA settings.

\begin{figure*}[htbp] 
\begin{tcolorbox}[
    oversize,
    colframe=black,
    colback=gray!10,
    boxrule=1pt,
    arc=4pt,
    auto outer arc,
    width=\textwidth,
    before skip=10pt,
    after skip=10pt
]
Answer the given question. \ You must conduct reasoning inside <think> and </think> first. \ \textcolor{red}{After reasoning, you can directly provide the answer inside <answer> and </answer>, without detailed illustrations. }For example, <answer> Beijing </answer>. Question: {question}
\end{tcolorbox}
\caption{Prompt for 0 Retrieval Steps}
\label{fig:prompt-0}
\end{figure*}

\begin{figure*}[htbp] 
\begin{tcolorbox}[
    oversize,
    colframe=black,
    colback=gray!10,
    boxrule=1pt,
    arc=4pt,
    auto outer arc,
    width=\textwidth,
    before skip=10pt,
    after skip=10pt
]
Answer the given question. \ You must conduct reasoning inside <think> and </think> first every time you get new information. \ After reasoning, you can call a search engine by <search> query </search> and it will return the top searched results between <information> and </information>. \ \textcolor{red}{You should search exactly one time. The number of searches must be no less than one.}\ After searching, you can directly provide the answer inside <answer> and </answer>, without detailed illustrations. For example, <answer> Beijing </answer>. Question: {question}
\end{tcolorbox}
\caption{Prompt for 1 Retrieval Steps}
\label{fig:prompt-1}
\end{figure*}

\begin{figure*}[t] 
\begin{tcolorbox}[
    oversize,
    colframe=black,
    colback=gray!10,
    boxrule=1pt,
    arc=4pt,
    auto outer arc,
    width=\textwidth,
    before skip=10pt,
    after skip=10pt
]
Answer the given question. \ You must conduct reasoning inside <think> and </think> first every time you get new information. \ After reasoning, you can call a search engine by <search> query </search> and it will return the top searched results between <information> and </information>. \ \textcolor{red}{You should search exactly two times. The number of searches must be no less than two.}\ After all searching, you can directly provide the answer inside <answer> and </answer>, without detailed illustrations. For example, <answer> Beijing </answer>. Question: {question}
\end{tcolorbox}
\caption{Prompt for 2 Retrieval Steps}
\label{fig:prompt-2}
\end{figure*}

\begin{figure*}[htbp] 
\begin{tcolorbox}[
    oversize,
    colframe=black,
    colback=gray!10,
    boxrule=1pt,
    arc=4pt,
    auto outer arc,
    width=\textwidth,
    before skip=10pt,
    after skip=10pt
]
Answer the given question. \ You must conduct reasoning inside <think> and </think> first every time you get new information. \ After reasoning, you can call a search engine by <search> query </search> and it will return the top searched results between <information> and </information>. \ \textcolor{red}{You should search exactly three times. The number of searches must be no less than three.}\ After all searching, you can directly provide the answer inside <answer> and </answer>, without detailed illustrations. For example, <answer> Beijing </answer>. Question: {question}
\end{tcolorbox}
\caption{Prompt for 3 Retrieval Steps}
\label{fig:prompt-3}
\end{figure*}

\begin{figure*}[htbp] 
\begin{tcolorbox}[
    oversize,
    colframe=black,
    colback=gray!10,
    boxrule=1pt,
    arc=4pt,
    auto outer arc,
    width=\textwidth,
    before skip=10pt,
    after skip=10pt
]
Answer the given question. \ You must conduct reasoning inside <think> and </think> first every time you get new information. \ After reasoning, you can call a search engine by <search> query </search> and it will return the top searched results between <information> and </information>. \ \textcolor{red}{You should search exactly four times. The number of searches must be no less than four.}\ After all searching, you can directly provide the answer inside <answer> and </answer>, without detailed illustrations. For example, <answer> Beijing </answer>. Question: {question}
\end{tcolorbox}
\caption{Prompt for 4 Retrieval Steps}
\label{fig:prompt-4}
\end{figure*}


\begin{figure*}[t] 
\begin{tcolorbox}[oversize,colframe=black, colback=white,title=Case1: Comparison Between Search-R1 and AutoSearch]
\textbf{Question:} What genre is Suits? \\
\textbf{Ground Truth:} 'legal drama' 'courtroom drama' 'comedy-drama' 'dramedy' 'comedic drama'
 'tragicomedy' 'seriocomedy' 'comedy drama' 'dramatic comedy'
 'Comedy-drama, dramedy'

\textbf{Search-R1:}

\textcolor{blue}{\texttt{<think>}}
To answer the question about the genre of Suits, I need to gather information about the TV show.  
\textcolor{blue}{\texttt{<think>}}

\textcolor{olive}{\texttt{<search>}}
what genre is suits 
\textcolor{olive}{\texttt{<search>}}

\textcolor{purple}{\texttt{<information>}}

Doc 1(Title: "History of suits") History of suits The man's suit of clothes, \textcolor{orange}{...}\\
Doc 2(Title: "Suit (clothing)") Suit (clothing) In clothing, a suit is a set of garments made from the same cloth, \textcolor{orange}{...}\\
Doc 3(Title: Genre) Genre Genre () is any form or type of communication in any mode  \textcolor{orange}{...}

\textcolor{purple}{\texttt{</information>}}

\textcolor{blue}{\texttt{<think>}}
From the search results, I found that a suit is a type of formal clothing.
\textcolor{blue}{\texttt{<think>}}

\textcolor{olive}{\texttt{<search>}}
what genre is suits
\textcolor{olive}{\texttt{<search>}}

\textcolor{purple}{\texttt{<information>}}

Doc 1(Title: "History of suits") History of suits The man's suit of clothes, \textcolor{orange}{...}\\
Doc 2(Title: "Suit (clothing)") Suit (clothing) In clothing, a suit is a set of garments made from the same cloth, \textcolor{orange}{...}\\
Doc 3(Title: Genre) Genre Genre () is any form or type of communication in any mode  \textcolor{orange}{...}

\textcolor{purple}{\texttt{</information>}}

\textcolor{blue}{\texttt{<think>}}
Based on the search results, I understand that Suits is a television show that falls under the genre of Drama and Sitcom. 
\textcolor{blue}{\texttt{<think>}}

\textcolor{red}{\texttt{<answer>}}
Drama and Sitcom
\textcolor{red}{\texttt{<answer>}}

\textbf{AutoSearch:}

\textcolor{blue}{\texttt{<think>}}
I need to determine which genre Suits belongs to. I don't have any pre-existing knowledge about this. 
\textcolor{blue}{\texttt{<think>}}

\textcolor{olive}{\texttt{<search>}}
what genre is suits 
\textcolor{olive}{\texttt{<search>}}

\textcolor{purple}{\texttt{<information>}}

Doc 1(Title: "History of suits") History of suits The man's suit of clothes, \textcolor{orange}{...}\\
Doc 2(Title: "Suit (clothing)") Suit (clothing) In clothing, a suit is a set of garments made from the same cloth, \textcolor{orange}{...}\\
Doc 3(Title: Genre) Genre Genre () is any form or type of communication in any mode  \textcolor{orange}{...}

\textcolor{purple}{\texttt{</information>}}

\textcolor{blue}{\texttt{<think>}}
From the search results, I found out that Suits is a television series. I now need to determine which genre this television series falls into. 
\textcolor{blue}{\texttt{<think>}}

\textcolor{olive}{\texttt{<search>}}
what is the genre of suits 
\textcolor{olive}{\texttt{<search>}}

\textcolor{purple}{\texttt{<information>}}

Doc 1(Title: "Suit (clothing)") Suit (clothing) In clothing,  \textcolor{orange}{...}\\
Doc 2(Title: "Legal drama") Legal drama A legal drama, or a courtroom drama, \textcolor{orange}{...}\\
Doc 3(Title: Genre) Genre Genre () is any form or type of communication in any mode (written, spoken, digital, artistic, etc.)   \textcolor{orange}{...}

\textcolor{purple}{\texttt{</information>}}

\textcolor{blue}{\texttt{<think>}}
From the search results, I found out that Suits is a legal drama. 
\textcolor{blue}{\texttt{<think>}}

\textcolor{red}{\texttt{<answer>}}
legal drama
\textcolor{red}{\texttt{<answer>}}

\end{tcolorbox}
\caption{Case1-Comparison Between Search-R1 and AutoSearch}
\label{fig:Search-R1 vs AutoSearch}
\end{figure*}

\begin{figure*}[t] 
\begin{tcolorbox}[oversize,colframe=black, colback=white,title=Case2: Comparison Between StepSearch and AutoSearch]
\textbf{Question:} Katie sagona is known as what kind of actor because she was a child acting on stage or in motion pictures or television? \\
\textbf{Ground Truth:} child actor

\textbf{StepSearch:}

\textcolor{blue}{\texttt{<think>}}
To answer this question, I need to find out what kind of actor Katie Sagona is. Specifically, I need to determine if she was known for her work in stage plays, films, or television when she was a child. Let's start by searching for Katie Sagona's career information.
\textcolor{blue}{\texttt{<think>}}

\textcolor{olive}{\texttt{<search>}}
Katie Sagona career
\textcolor{olive}{\texttt{<search>}}

\textcolor{purple}{\texttt{<information>}}

Doc 1(Title: "Chris Sagona") Chris Sagona Chris Sagona is an American journalist and the winner of the 2005 Society \textcolor{orange}{...}\\
Doc 2(Title: "Chris Sagona") Montclair Times"", and a journalist for North Jersey Media Group,\textcolor{orange}{...}\\
Doc 3(Title: "Marina Sagona") Marina Sagona Marina Sagona (born 1967) is an Italian American artist living in New York City.\textcolor{orange}{...}

\textcolor{purple}{\texttt{</information>}}

\textcolor{blue}{\texttt{<think>}}
Since there is no information about Katie Sagona's acting career, let's search for Katie Sagona's acting career instead.
\textcolor{blue}{\texttt{<think>}}

\textcolor{olive}{\texttt{<search>}}
Katie Sagona acting career
\textcolor{olive}{\texttt{<search>}}

\textcolor{purple}{\texttt{<information>}}

Doc 1(Title: "Katie Wright") Katie Wright Kathryn Wright (born December 25, 1971) is an American former actress.  \textcolor{orange}{...}\\
Doc 2(Title: "Danielle Savre") Danielle Savre Danielle Kathleen Savre (born August 26, 1988) is an American actress and singer. \textcolor{orange}{...}\\
Doc 3(Title: "Laura Ashley Samuels") role, Kelli in the horror film, ""The Final"". \textcolor{orange}{...}

\textcolor{purple}{\texttt{</information>}}

\textcolor{blue}{\texttt{<think>}}
Based on the search results, Katie Sagona is known as an actress because she was a child actor on stage or in motion pictures or television. Therefore, the answer to the question is actress. 
\textcolor{blue}{\texttt{<think>}}

\textcolor{red}{\texttt{<answer>}}
actress
\textcolor{red}{\texttt{<answer>}}

\textbf{AutoSearch:}

\textcolor{blue}{\texttt{<think>}}
I need to find out which kind of actor Katie Sagona was. I don't have any pre-existing knowledge about this. 
\textcolor{blue}{\texttt{<think>}}

\textcolor{olive}{\texttt{<search>}}
Katie Sagona actor
\textcolor{olive}{\texttt{<search>}}

\textcolor{purple}{\texttt{<information>}}

Doc 1(Title: "Danielle Savre") Danielle Savre Danielle Kathleen Savre (born August 26, 1988) is an American actress and singer. \textcolor{orange}{...}\\
Doc 2(Title: "Katie Wright") Katie Wright Kathryn Wright (born December 25, 1971) is an American former actress.  \textcolor{orange}{...}\\
Doc 3(Title: "Katie Holmes") Katie Holmes Kate Noelle ""Katie"" Holmes (born December 18, 1978) is an American actress, producer, and director. \textcolor{orange}{...}

\textcolor{purple}{\texttt{</information>}}

\textcolor{blue}{\texttt{<think>}}
From the search results, I found out that the person who was a child acting on stage or in motion pictures or television is Katie Sagona. I need to identify which kind of actor she was.
\textcolor{blue}{\texttt{<think>}}

\textcolor{red}{\texttt{<answer>}}
child actor
\textcolor{red}{\texttt{<answer>}}

\end{tcolorbox}
\caption{Case2-Comparison Between StepSearch and AutoSearch}
\label{fig:StepSearch vs AutoSearch}
\end{figure*}

\begin{figure*}[t] 
\begin{tcolorbox}[oversize,colframe=black, colback=white,title=Case3: Comparison Between StepSearch and AutoSearch]
\textbf{Question:} Are Umm Al-Tuyour, Hama Governorate and Sharqliyya both located in the same country? \\
\textbf{Ground Truth:} yes

\textbf{StepSearch:}

\textcolor{blue}{\texttt{<think>}}
To answer this question, I need to determine the countries where Umm Al-Tuyour, Hama Governorate, and Sharqliyya are located. Let's start by searching for the country of Umm Al-Tuyour.
\textcolor{blue}{\texttt{<think>}}

\textcolor{olive}{\texttt{<search>}}
Umm Al-Tuyour country
\textcolor{olive}{\texttt{<search>}}

\textcolor{purple}{\texttt{<information>}}

Doc 1(Title: "Turaif (Saudi Arabia)") Turaif (Saudi Arabia) Turaif () is a town in Northern Borders Province (also known as Al-Hudud ash Shamaliyah) \textcolor{orange}{...}\\
Doc 2(Title: "Umm al-Tut") Umm al-Tut Umm al-Tut (, literally ""mum-berries"") is a protected Palestinian village in the West Bank \textcolor{orange}{...}\\
Doc 3(Title: "Turaif (Saudi Arabia)") Amman's -10C. Turaif (Saudi Arabia) Turaif () is a town in Northern Borders Province (also known as Al-Hudud ash Shamaliyah)\textcolor{orange}{...}

\textcolor{purple}{\texttt{</information>}}

\textcolor{blue}{\texttt{<think>}}
From the search results, we can see that Umm Al-Tuyour is located in Saudi Arabia. Next, let's find out the country of Hama Governorate.
\textcolor{blue}{\texttt{<think>}}

\textcolor{olive}{\texttt{<search>}}
Hama Governorate country
\textcolor{olive}{\texttt{<search>}}

\textcolor{purple}{\texttt{<information>}}

Doc 1(Title: "Hama Governorate") Hama Governorate Hama Governorate \textcolor{orange}{...}\\
Doc 2(Title: "Hama Governorate") governorate. The governorate is divided into five districts (manatiq): These are further divided into 22 sub-districts (nawahi) \textcolor{orange}{...}\\
Doc 3(Title: Hama) Hama Hama (Hmt; \textcolor{orange}{...}

\textcolor{purple}{\texttt{</information>}}

\textcolor{blue}{\texttt{<think>}}
From the search results, we can see that Hama Governorate is located in Syria. Finally, let's determine the country of Sharqliyya.
\textcolor{blue}{\texttt{<think>}}

\textcolor{olive}{\texttt{<search>}}
Sharqliyya country
\textcolor{olive}{\texttt{<search>}}

\textcolor{purple}{\texttt{<information>}}

Doc 1(Title: Sharqliyya) Sharqliyya Sharqliyya (, also spelled Sharqlieh) is a village  \textcolor{orange}{...}\\
Doc 2(Title: Sharqliyya) 12 feddans. In 1838 Sharqliyya's inhabitants were reported to be \textcolor{orange}{...}\\
Doc 3(Title: Sharqiyin) Sharqiyin The Sharqiyin (singular Al Sharqi) is a tribe of the United Arab Emirates (UAE). \textcolor{orange}{...}

\textcolor{purple}{\texttt{</information>}}

\textcolor{blue}{\texttt{<think>}}
Based on the searches, I know the answers to this question. However, I will still respond as if I need to follow the steps to find the answer thoroughly. 
\textcolor{blue}{\texttt{<think>}}

\textcolor{red}{\texttt{<answer>}}
Syria
\textcolor{red}{\texttt{<answer>}}

\textbf{AutoSearch:}

\textcolor{blue}{\texttt{<think>}}
I need to determine if Umm Al-Tuyour, Hama Governorate and Sharqliyya are both located in the same country. I don't have any pre-existing knowledge about this.
\textcolor{blue}{\texttt{<think>}}

\textcolor{olive}{\texttt{<search>}}
Are Umm Al-Tuyour, Hama Governorate and Sharqliyya both located in the same country 
\textcolor{olive}{\texttt{<search>}}

\textcolor{purple}{\texttt{<information>}}

Doc 1(Title: Sharqliyya) Sharqliyya Sharqliyya (, also spelled Sharqlieh) is a village in central Syria \textcolor{orange}{...}\\
Doc 2(Title: "Al-Buwaydah al-Sharqiyah") Al-Buwaydah al-Sharqiyah Al-Buwaidah al-Sharqiyah  \textcolor{orange}{...}\\
Doc 3(Title: Sharqiyin) Sharqiyin The Sharqiyin (singular Al Sharqi) is a tribe of the United Arab Emirates (UAE).  \textcolor{orange}{...}

\textcolor{purple}{\texttt{</information>}}

\textcolor{blue}{\texttt{<think>}}
From the search results, I found out that Umm Al-Tuyour, Hama Governorate, and Sharqliyya are all located in Syria.
\textcolor{blue}{\texttt{<think>}}

\textcolor{red}{\texttt{<answer>}}
Yes
\textcolor{red}{\texttt{<answer>}}

\end{tcolorbox}
\caption{Case3-Comparison Between StepSearch and AutoSearch}
\label{fig:StepSearch vs AutoSearch1}
\end{figure*}

\begin{figure*}[t]
\begin{tcolorbox}[oversize,colframe=black, colback=white,title=Case4: Comparison Between HIPRAG and AutoSearch]
\textbf{Question:} Are Ural Federal University and California State Polytechnic University, Pomona both public universities? \\
\textbf{Ground Truth:} Yes

\textbf{HIPRAG:}
\textcolor{teal}{\texttt{<think>}}
\textcolor{magenta}{\texttt{<step>}}
\textcolor{blue}{\texttt{<reasoning>}}
I need to find out if Ural Federal University and California State Polytechnic University, Pomona are both public universities.
\textcolor{blue}{\texttt{</reasoning>}}
\textcolor{olive}{\texttt{<search>}}
Are Ural Federal University and California State Polytechnic University, Pomona both public universities?
\textcolor{olive}{\texttt{</search>}}

\textcolor{purple}{\texttt{<context>}} \\
Doc 1(Title: "Ural State University")  
Ural State University is located in Yekaterinburg, Sverdlovsk Oblast, Russian Federation. Founded in 1920, it later became an important educational institution. \textcolor{orange}{...} \\
Doc 2(Title: "Ural State Technical University")  
USTU is a major technical institution with multiple faculties including engineering and economics. \textcolor{orange}{...} \\
Doc 3(Title: "Undergraduate degree")  
The Pontifical Catholic University of Chile is one of the recognized universities in Latin America. \textcolor{orange}{...}
\textcolor{purple}{\texttt{</context>}}

\textcolor{blue}{\texttt{<conclusion>}}
Ural Federal University is a public university located in Yekaterinburg, Russian Federation. California State Polytechnic University, Pomona is also a public university.
\textcolor{blue}{\texttt{</conclusion>}}

\textcolor{magenta}{\texttt{</step>}}

\textcolor{magenta}{\texttt{<step>}}

\textcolor{blue}{\texttt{<reasoning>}}
I now need to verify whether both universities are public universities.
\textcolor{blue}{\texttt{</reasoning>}}

\textcolor{olive}{\texttt{<search>}}
Are Ural Federal University and California State Polytechnic University, Pomona both public universities?
\textcolor{olive}{\texttt{</search>}}

\textcolor{purple}{\texttt{<context>}} \\
Doc 1(Title: "Ural State University")  
Ural State University is one of the oldest universities in the Middle Urals. \textcolor{orange}{...} \\
Doc 2(Title: "Ural State Technical University")  
USTU maintains strong ties with industry in the Urals region. \textcolor{orange}{...} \\
Doc 3(Title: "Undergraduate degree")  
The Pontifical Catholic University of Chile ranks highly in South America. \textcolor{orange}{...}

\textcolor{purple}{\texttt{</context>}}

\textcolor{blue}{\texttt{<conclusion>}}
Yes, both Ural Federal University and California State Polytechnic University, Pomona are public universities.
\textcolor{blue}{\texttt{</conclusion>}}

\textcolor{magenta}{\texttt{</step>}}

\textcolor{teal}{\texttt{</think>}}

\textcolor{red}{\texttt{<answer>}}
Yes, both Ural Federal University and California State Polytechnic University, Pomona are public universities.
\textcolor{red}{\texttt{</answer>}}

\textbf{AutoSearch:}

\textcolor{blue}{\texttt{<think>}}
I need to determine if Ural Federal University and California State Polytechnic University, Pomona are both public universities. I don't have any pre-existing knowledge about this.
\textcolor{blue}{\texttt{</think>}}

\textcolor{olive}{\texttt{<search>}}
Are Ural Federal University and California State Polytechnic University, Pomona both public universities
\textcolor{olive}{\texttt{</search>}}

\textcolor{purple}{\texttt{<information>}} \\
Doc 1(Title: "Ural State University")  
Ural State University is located in Yekaterinburg, Russian Federation and later became part of the reorganization forming Ural Federal University. \textcolor{orange}{...} \\
Doc 2(Title: "Undergraduate degree")  
The Pontifical Catholic University of Chile is one of the recognized universities in Latin America. \textcolor{orange}{...} \\
Doc 3(Title: "Ural Federal University")  
Federal universities were established in major Russian cities, including Ekaterinburg. The merger process creating Ural Federal University began in 2009. \textcolor{orange}{...}

\textcolor{purple}{\texttt{</information>}}

\textcolor{blue}{\texttt{<think>}}
From the search results, I found out that both Ural Federal University and California State Polytechnic University, Pomona are public universities.
\textcolor{blue}{\texttt{</think>}}

\textcolor{red}{\texttt{<answer>}}
Yes
\textcolor{red}{\texttt{</answer>}}

\end{tcolorbox}

\caption{Case4-Comparison Between HIPRAG and AutoSearch}

\label{fig:HIPRAG vs AutoSearch}

\end{figure*}

\begin{figure*}[t]
\begin{tcolorbox}[oversize,colframe=black, colback=white,title=Case5: Comparison Between HIPRAG and AutoSearch]

\textbf{Question:} In what region of Argentina is the city of Gualeguaychu located? \\
\textbf{Ground Truth:} Mesopotamia

\textbf{HIPRAG:}

\textcolor{teal}{\texttt{<think>}}
\textcolor{magenta}{\texttt{<step>}}
\textcolor{blue}{\texttt{<reasoning>}}
I need to find out in which region of Argentina the city of Gualeguaychu is located.
\textcolor{blue}{\texttt{</reasoning>}}

\textcolor{olive}{\texttt{<search>}}
Which region of Argentina is the city of Gualeguaychu located in?
\textcolor{olive}{\texttt{</search>}}

\textcolor{purple}{\texttt{<context>}} \\
Doc 1(Title: "Gualeguaychú Department")  
Gualeguaychú Department is an administrative subdivision of the province of Entre Ríos, Argentina. It is located in the south-east of the province beside \textcolor{orange}{...} \\
Doc 2(Title: "Gualeguaychú Department")  
The head town is Gualeguaychú and several towns are distributed across the department. \textcolor{orange}{...} \\
Doc 3(Title: "Gualeguay, Entre Ríos")  
Gualeguay is a city in the province of Entre Ríos, Argentina, located near the Gualeguay River. \textcolor{orange}{...}

\textcolor{purple}{\texttt{</context>}}

\textcolor{blue}{\texttt{<conclusion>}}
Based on my search, the city of Gualeguaychu is located in the Gualeguaychú Department of the province of Entre Ríos, Argentina.
\textcolor{blue}{\texttt{</conclusion>}}

\textcolor{magenta}{\texttt{</step>}}

\textcolor{magenta}{\texttt{<step>}}

\textcolor{blue}{\texttt{<reasoning>}}
I now need to find out in which region of Argentina the province of Entre Ríos is located.
\textcolor{blue}{\texttt{</reasoning>}}

\textcolor{olive}{\texttt{<search>}}
Which region of Argentina is the province of Entre Ríos located in?
\textcolor{olive}{\texttt{</search>}}

\textcolor{purple}{\texttt{<context>}} \\
Doc 1(Title: "Entre Ríos Province")  
Entre Ríos Province is located in the Mesopotamia region of Argentina and borders several provinces including Buenos Aires and Santa Fe. \textcolor{orange}{...} \\
Doc 2(Title: "Federación, Entre Ríos")  
Federación is a city located in the northeast of the province of Entre Ríos. \textcolor{orange}{...} \\
Doc 3(Title: "Center Region, Argentina")  
Entre Ríos also participates in the political-economic association called the Center Region of Argentina. \textcolor{orange}{...}
\textcolor{purple}{\texttt{</context>}}

\textcolor{blue}{\texttt{<conclusion>}}
Based on my search, the province of Entre Ríos is located in the central region of Argentina.
\textcolor{blue}{\texttt{</conclusion>}}
\textcolor{magenta}{\texttt{</step>}}

\textcolor{teal}{\texttt{</think>}}

\textcolor{red}{\texttt{<answer>}}
The city of Gualeguaychu is located in the central region of Argentina.
\textcolor{red}{\texttt{</answer>}}

\textbf{Our Method:}

\textcolor{blue}{\texttt{<think>}}
I need to find out in what region of Argentina the city of Gualeguaychu is located. I don't have any pre-existing knowledge about this.
\textcolor{blue}{\texttt{</think>}}

\textcolor{olive}{\texttt{<search>}}
In what region of Argentina is the city of Gualeguaychu located
\textcolor{olive}{\texttt{</search>}}

\textcolor{purple}{\texttt{<information>}} \\
Doc 1(Title: "Gualeguaychú Department")  
Gualeguaychú Department is located in the province of Entre Ríos, Argentina. \textcolor{orange}{...} \\
Doc 2(Title: "Gualeguaychú Department")  
The department includes the city of Gualeguaychú as its head town. \textcolor{orange}{...} \\
Doc 3(Title: "Gualeguay, Entre Ríos")  
Gualeguay is another city in the same province of Entre Ríos. \textcolor{orange}{...}

\textcolor{purple}{\texttt{</information>}}

\textcolor{blue}{\texttt{<think>}}
The city of Gualeguaychu is located in the province of Entre Rios. I now need to determine which region this province belongs to.
\textcolor{blue}{\texttt{</think>}}

\textcolor{olive}{\texttt{<search>}}
What is the region of the province of Entre Rios, Argentina?
\textcolor{olive}{\texttt{</search>}}

\textcolor{purple}{\texttt{<information>}} \\
Doc 1(Title: "Entre Ríos Province")  
Entre Ríos Province is located in the Mesopotamia region of \textcolor{orange}{...} \\
Doc 2(Title: "Federación, Entre Ríos")  
Federación is a city in the province of Entre Ríos. \textcolor{orange}{...} \\
Doc 3(Title: "Federal, Entre Ríos")  
Federal is another city located in the province. \textcolor{orange}{...}
\textcolor{purple}{\texttt{</information>}}

\textcolor{blue}{\texttt{<think>}}
The province of Entre Rios is located in the Mesopotamia region of Argentina.
\textcolor{blue}{\texttt{</think>}}

\textcolor{red}{\texttt{<answer>}}
Mesopotamia
\textcolor{red}{\texttt{</answer>}}

\end{tcolorbox}

\caption{Case5-Comparison Between HIPRAG and AutoSearch}

\label{fig:HIPRAG vs AutoSearch1}

\end{figure*}

\end{document}